\documentclass[10pt]{article} 

\usepackage[preprint]{tmlr}

\usepackage[utf8]{inputenc}
\usepackage{amsmath}
\usepackage{amssymb}
\usepackage{amsfonts}
\usepackage{mathtools}
\usepackage{amsthm}
\usepackage{bm}
\usepackage{graphicx}
\usepackage{wrapfig}
\usepackage{subcaption}
\usepackage{booktabs}
\usepackage{makecell}
\usepackage{multirow}
\usepackage{nicefrac}
\usepackage{microtype}
\usepackage{xcolor}
\usepackage{url}
\usepackage{hyperref}
\usepackage[capitalize,noabbrev]{cleveref}

\theoremstyle{plain}

\theoremstyle{definition}

\theoremstyle{remark}

\title{L2R: Low-Rank and Lipschitz-Controlled Routing\\ for Mixture-of-Experts }

\author{
  \name\hspace{-1ex}Minghao Yang \quad Ren Togo \quad Guang Li\thanks{Correspondence to Guang Li \texttt{<guang@lmd.ist.hokudai.ac.jp>}.} \quad Takahiro Ogawa \quad Miki Haseyama \\
  \addr Hokkaido University \\
  \texttt{\{yang, togo, guang, ogawa, mhaseyama\}@lmd.ist.hokudai.ac.jp}
}

\begin{document}

\maketitle

\begin{abstract}
Mixture-of-Experts (MoE) models scale neural networks by conditionally activating a small subset of experts, where the router plays a central role in determining expert specialization and overall model performance.
However, many modern MoE systems still adopt linear routers in raw high-dimensional representation spaces, where representation mismatch, angular concentration, and scale-sensitive scoring can jointly undermine routing discriminability and stable expert specialization.
In this work, we propose Low-rank \& Lipschitz-controlled Routing (L2R), a unified routing framework that reshapes both the routing space and scoring geometry. L2R performs expert assignment in a shared low-rank latent routing space and introduces Saturated Inner-Product Scoring (SIPS) to explicitly control the Lipschitz behavior of routing functions, yielding smoother and more stable routing geometry. In addition, L2R incorporates a parameter-efficient multi-anchor routing mechanism to enhance expert expressiveness. Experiments on an OLMoE-based language MoE model and a ViT-based ImageNet setting show improved overall performance in both domains; OLMoE diagnostics further show improved routing geometry and expert discrimination.
\end{abstract}

\newcommand{\method}{{L2R}}
\newcommand{\methodfull}{{Low-rank \& Lipschitz-controlled Routing}}

\section{Introduction}
Mixture-of-Experts (MoE) has emerged as a leading paradigm for scaling deep neural networks across a wide range of domains, including large language models (LLMs)~\citep{muennighoff2025olmoe,deepseekmoe,jiang2024mixtral,yang2024qwen2} and vision backbones~\citep{vmoe,chen2023modsquad,yang2025ase}. By enabling conditional computation, MoE decouples model capacity from inference cost and is designed to realize meaningful expert specialization~\citep{cai2025survey}. Central to achieving this objective is the router. As the mechanism that assigns tokens to experts under sparse and dynamic activation, the router effectively defines the functional topology of the model, governing how capacity is allocated and coordinated. Consequently, the quality of routing strongly influences the expressiveness, stability, and effective capacity of MoE models~\citep{dikkala2023routing}.

Despite these advances, modern MoE routing still exhibits two practical limitations: 
1) \textbf{Raw-space routing and high-dimensional concentration:} Routing is typically performed in raw, high-dimensional representation spaces that are optimized for representation learning rather than expert discrimination, resulting in a fundamental representation mismatch. Moreover, in high-dimensional spaces, similarities tend to concentrate~\citep{cai2013angle}, reducing effective separability among experts~\citep{chi2022on}.
2) \textbf{Scale-sensitive dot-product scoring:} Standard dot-product logits entangle directional alignment with feature scale, making routing highly sensitive to magnitude variations and encouraging norm-dominated competition among expert anchors, which introduces ambiguity in expert selection and weakens expert discrimination.

\begin{figure}[t]
  \centering
  \includegraphics[width=\textwidth]{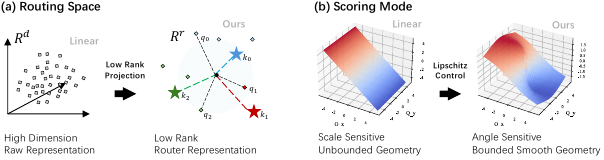}
  \caption{\textbf{Comparison between the linear router (left) and the proposed L2R framework (right).}
  (a) Routing space: Tokens are projected from high-dimensional raw representations into a low-rank routing space as $\bm{q}$, where experts are represented by learnable anchors $\bm{k}$.
  (b) Scoring mode: Compared to dot-product scoring, SIPS reshapes the score landscape into a bounded and smoother geometry, improving routing stability and expert discrimination.}
  \label{overview}
  \vspace{-2em}
\end{figure}

To address these issues, we propose \textbf{L}ow-rank \& \textbf{L}ipschitz-controlled \textbf{R}outing (\textbf{L2R}), a unified routing framework with three components: a low-rank routing space for representation decoupling, a Lipschitz-controlled scoring function for stable expert selection, and a parameter-efficient multi-anchor mechanism for enhanced expert expressiveness. 
\textbf{First}, instead of performing expert selection directly in the raw high-dimensional representation space, tokens are mapped into a shared low-rank routing space that produces routing-specific queries. Experts are represented by learnable anchors in this latent space, and routing is performed by matching queries to anchors. This design decouples routing from representation learning, compresses the search space for expert positioning, and alleviates geometric concentration in high-dimensional routing. 
\textbf{Second}, Saturated Inner-Product Scoring (SIPS) is introduced to bound the sensitivity of routing scores to representation scale and perturbations, thereby explicitly controlling the Lipschitz behavior of the routing function. By reshaping the score landscape into a smoother and more stable geometry, SIPS reduces routing ambiguity and promotes robust expert discrimination. 
\textbf{Finally}, we develop a parameter-efficient multi-anchor routing mechanism that associates each expert with multiple anchors, allowing a single expert to capture diverse semantic views without incurring prohibitive overhead.

We extensively evaluate the proposed L2R framework across both language and vision MoE settings to assess its generality and effectiveness. Specifically, we conduct from-scratch training on large-scale language MoE models based on OLMoE~\citep{muennighoff2025olmoe} and evaluate vision MoE architectures on ImageNet using ViT backbones. Across both domains, our method consistently improves overall model performance, while OLMoE diagnostics further show improved routing geometry and expert discrimination, validating its effectiveness in both language and vision domains.
Our contributions are summarized as follows:
\begin{itemize}
\item We identify two practical limitations of linear MoE routing, namely routing in raw high-dimensional representation spaces and scale-sensitive dot-product scoring, which lead to routing ambiguity and insufficient expert specialization.
\item We propose Low-rank \& Lipschitz-controlled Routing (L2R), a unified routing framework that performs expert assignment in a low-rank latent routing space with Saturated Inner-Product Scoring (SIPS) and parameter-efficient multi-anchor routing, yielding smoother, more stable, and more expressive routing geometry.
\item We evaluate L2R across both language and vision MoE settings, including OLMoE-based language models and vision backbones on ImageNet, showing improved overall performance in both settings and improved routing geometry and expert discrimination in OLMoE diagnostics.
\end{itemize}

\section{Related Work}
\label{related_works}
\paragraph{Router Design in Sparse MoE.}
Most modern sparse MoE systems still rely on {raw-space linear routing}, where expert logits are produced by a single linear projection in the backbone representation space, e.g., in OLMoE~\citep{muennighoff2025olmoe}, Mixtral~\citep{jiang2024mixtral}, Qwen2-MoE~\citep{yang2024qwen2}, and DeepSeek-MoE~\citep{deepseekmoe}, as well as vision MoE variants such as V-MoE~\citep{vmoe}, Mod-Squad~\citep{chen2023modsquad}, and ASE~\citep{yang2025ase}.
Recent studies revisit router design from multiple angles: improving routing expressiveness (e.g., Router Upcycling~\citep{ran2025routerupcycling} and attention-based routing in Yuan~2.0-M32~\citep{wu2024yuan20m32mixtureexperts}), and improving stability and utilization via auxiliary objectives and routing variants~\citep{shazeer2017outrageously,lepikhin2020gshard,fedus2021switch}.
Despite these advances, the routing score is still typically produced by \emph{linear scoring in a high-dimensional space}, which leaves the scoring geometry sensitive to scale and high-dimensional concentration.

\paragraph{Geometry-Aware Latent-Space Routing.}
Another line of work performs routing in a \emph{latent space} and adopts angle-aware similarities~\citep{chi2022on,nguyen2024statistical,wu2024multihead}.
X-MoE~\citep{chi2022on} projects representations to a lower-dimensional space and uses cosine scoring to alleviate collapse in high-dimensional gating.
Different from prior works, our study emphasizes two structural limitations that jointly weaken discriminability under raw-space routing: representation mismatch and angular concentration.
We further explore routing rank across language and vision settings, and find that a small rank can provide competitive performance across domains, motivating the use of a consistent low-rank configuration across settings.
Within this latent-space routing paradigm, \method{} introduces \emph{SIPS}, a Lipschitz-controlled scoring geometry that retains magnitude cues while explicitly bounding their effective influence (instead of discarding magnitude as cosine scoring), and improves expressiveness via lightweight multi-anchor heads.

\section{Methodology}
In this section, we first introduce the standard formulation of MoE architecture and the commonly adopted linear routing mechanism. We then present the proposed L2R framework that reshapes both the routing space and the scoring geometry for stable and expressive expert specialization.

\subsection{Preliminaries}
\label{sec:preliminaries}

\paragraph{Mixture-of-Experts.}
MoE models scale network capacity by replicating a sub-module, such as the feed-forward network (FFN) in a Transformer block, into a set of $N$ parallel experts and activating only a sparse subset per input token~\citep{cai2025survey}.
Given an input representation $\bm{x} \in \mathbb{R}^d$, a router $g(\cdot)$ first produces routing logits:
\begin{equation}
    \bm{z} = g(\bm{x}) \in \mathbb{R}^N,
\end{equation}
where each element $z_i$ reflects the affinity between the token and expert $E_i$.
The logits are then normalized via a softmax function with temperature $\tau$ to obtain routing scores:
\begin{equation}
    \bm{s} = \mathrm{Softmax}\!\left(\bm{z} / \tau \right) \in \mathbb{R}^N.
\end{equation}
A sparse gating mechanism subsequently selects the top-$k$ experts, and the final MoE output is computed as:
\begin{equation}
    \bm{y} = \bm{x} + \sum_{i \in \mathcal{T}} s_i E_i(\bm{x}), 
    \quad \mathcal{T} = \mathrm{TopK}(\bm{s}, k),
\end{equation}
where $\mathcal{T}$ denotes the indices of the selected experts.
The routing scores jointly determine expert activation and their relative contributions, thereby defining the effective mixture geometry and influencing expert specialization~\citep{dikkala2023routing}.

\paragraph{Linear Router.}
In most existing MoE architectures~\citep{muennighoff2025olmoe,jiang2024mixtral,yang2024qwen2,deepseekmoe}, the router $g(\cdot)$ is implemented as a dense linear transformation:
\begin{equation} \label{eql:linear} g(\bm{x}) = \bm{x} \bm{W}_g, \quad  \bm{W}_g = [\bm{w}_1, \ldots, \bm{w}_N] \in \mathbb{R}^{d \times N},
\end{equation}
where each column $\bm{w}_i$ serves as the geometric anchor of expert $E_i$.
The routing logit for expert $E_i$ is computed as:
\begin{equation}
    z_i = \bm{x}^\top \bm{w}_i,
\end{equation}
which induces a partition of the high-dimensional representation space $\mathbb{R}^d$ by linear decision hyperplanes defined by $\{\bm{w}_i\}_{i=1}^N$. This formulation performs expert assignment directly in raw backbone feature space using unbounded dot-product scoring, and thus inherits representation mismatch and scale-sensitive routing behaviors.

\subsection{Low-Rank Routing}
\label{sec:lowrank}

Linear routers perform expert assignment directly in the raw, high-dimensional backbone representation space. This design induces two structural limitations: 1) \textit{representation mismatch} and 2) \textit{high-dimensional concentration}, which limit effective expert specialization.

\paragraph{Representation Mismatch.}
Backbone representations $\bm{x}$ are primarily optimized for representation learning and downstream task prediction, rather than for expert discrimination.
Consequently, the router is forced to operate in a \emph{passive} feature space that is not explicitly shaped for separating tokens into expert-specific clusters.
Especially in LLMs, semantic factors in such representations are often highly entangled due to superposition effects~\citep{elhage2022toy}, and token distributions are not encouraged to organize into well-separated expert-aware regions.
This representation mismatch makes it difficult for a linear router to induce stable and semantically coherent expert assignments. 

\begin{wrapfigure}{r}{0.42\textwidth}
    \centering
    \vspace{-1.6em}
    \includegraphics[width=0.95\linewidth]{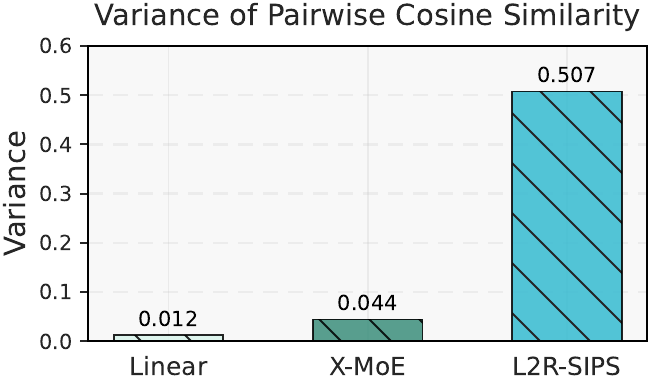}\hspace{1mm}   \vspace{-1mm}
       \caption{Layer-averaged variance of pairwise token cosine similarities in the routing space.}
    \label{fig:concentration}
    \vspace{-3mm}
\end{wrapfigure}

\paragraph{High-Dimensional Concentration.}
Linear routing typically operates in a high-dimensional regime with $d \gg N$ (e.g., $d=4,096$ with $N \leq 64$)~\citep{cai2025survey}. 
In such spaces, angular similarities tend to concentrate~\citep{cai2013angle}, causing random or weakly structured vectors to become nearly orthogonal. Figure~\ref{fig:concentration} compares the variance of pairwise token cosine similarities in the latent routing spaces of Linear, X-MoE~\citep{chi2022on}, and L2R-SIPS on OLMoE~\citep{muennighoff2025olmoe}.
It provides direct empirical evidence of this phenomenon:  backbone token representations for Linear routers exhibit near-zero variance in pairwise cosine similarity, indicating a highly concentrated angular geometry. As a result, routing logits across experts have poorly separated margins, leading to ambiguous expert selection.
Moreover, the large dimensionality of the routing space significantly
enlarges the search space for expert anchors, making anchor optimization
more difficult and hindering effective expert positioning.

\paragraph{Low-Rank Latent Routing Space.}
To address the above limitations, we introduce a shared low-rank latent routing space that explicitly decouples routing from raw backbone geometry.
Specifically, we parameterize routing through a shared projection:
\begin{equation}
    \bm{q} = f(\bm{x}) = \bm{x}\bm{W}_q, \quad \bm{W}_q \in \mathbb{R}^{d \times r}, \quad r \ll d,
\end{equation}
where $\bm{q} \in \mathbb{R}^{r}$ denotes the routing-specific query representation.
Each expert $E_i$ is associated with a learnable anchor $\bm{k}_i \in \mathbb{R}^{r}$ in this latent routing space, and routing logits are computed by matching $\bm{q}$ with $\{\bm{k}_i\}$.  This can be viewed as an attention-like operation, where $\bm{q}$ serves as a routing query and $\bm{k}_i$ as expert keys.

This formulation yields two immediate benefits:
First, the shared projection $\bm{W}_q$ explicitly learns a routing-oriented subspace, transforming backbone features into a space tailored for expert discrimination and {mitigating raw-space representation mismatch}.
Second, routing in the compressed space $\mathbb{R}^{r}$ substantially reduces geometric concentration and the anchor search space, enabling more stable anchor positioning and more discriminative expert selection. 
Figure~\ref{fig:concentration} empirically supports this effect: with rank $r{=}2$, L2R-SIPS yields much higher pairwise-cosine variance than Linear and X-MoE, approaching the isotropic 2D reference value of $0.5$ and indicating improved angular diversity.

\subsection{Saturated Inner-Product Scoring (SIPS)}
\label{sec:sips}

In this section, we analyze the limitations of standard dot-product scoring and introduce {Saturated Inner-Product Scoring (SIPS)}, a bounded scoring geometry that balances the contributions of magnitude and angle for stable and expressive expert selection.

\paragraph{Limitations of Dot-Product Scoring.}
With low-rank routing (Section~\ref{sec:lowrank}), standard dot-product scoring typically computes logits as:
\begin{equation}
    z_i = \bm{q}^\top \bm{k}_i
    = \|\bm{q}\|\,\|\bm{k}_i\| \cos\theta(\bm{q},\bm{k}_i).
\end{equation}
This coupling entangles \emph{directional alignment} with \emph{feature scale}.
In practice, (i) token magnitudes $\|\bm{q}\|$ can vary substantially (especially in LLMs, where certain token patterns yield unusually large activations), which scales all logits and effectively induces an input-dependent softmax temperature; and (ii) anchor norms $\|\bm{k}_i\|$ modulate inter-expert logit gaps and can dominate the competition, since increasing $\|\bm{k}_i\|$ often provides an easier route to raising scores than improving angular alignment.
Together, these effects make routing scale-sensitive and encourage norm-dominated expert competition, introducing ambiguity in expert selection and weakening expert specialization.

\begin{wrapfigure}{r}{0.44\textwidth}
    \vspace{-5mm}
    \centering
    \begin{subfigure}[t]{0.475\linewidth}
        \centering
        \includegraphics[width=\linewidth]{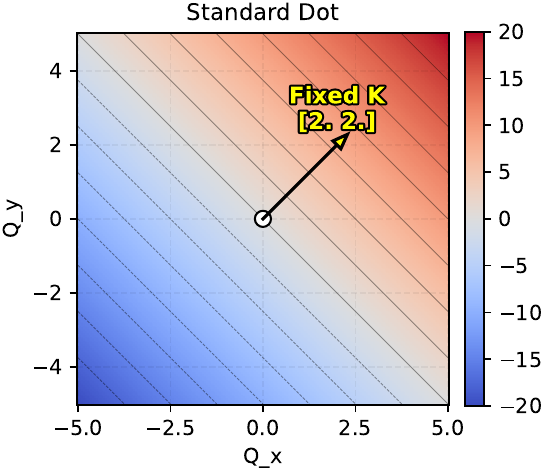}
        \caption{Dot Product Scoring}
        \label{fig:sips_heatmap_dot}
    \end{subfigure}
    \hfill
    \begin{subfigure}[t]{0.475\linewidth}
        \centering
        \includegraphics[width=\linewidth]{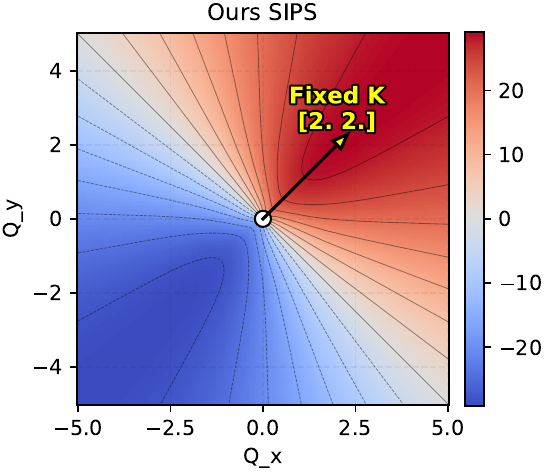}
        \caption{SIPS Scoring}
        \label{fig:sips_heatmap_sips}
    \end{subfigure}
    \vspace{-1mm}
    \caption{\textbf{Score landscapes under fixed expert anchor.}
    Heatmaps visualize routing logits as a function of query location $\bm{q}=(Q_x,Q_y)$ with a fixed anchor $\bm{k}=[2,2]$.
    }
    \label{fig:sips_heatmaps}
    \vspace{-7mm}
\end{wrapfigure}

\paragraph{Geometric Visualization.}
Figure~\ref{fig:sips_heatmaps} visualizes the logit landscape by fixing an expert anchor $\bm{k}$ and sweeping query locations $\bm{q}$ in a 2D routing plane.
Under standard dot-product scoring (Figure~\ref{fig:sips_heatmap_dot}), level sets are nearly parallel lines, and the score grows linearly with $\|\bm{q}\|$, hence a large region shares similar scores, yielding weak angular discrimination and making expert selection sensitive to magnitude variations.
This motivates a bounded scoring geometry that sharpens directional preference while controlling radial score growth, which we formalize next as SIPS.

\paragraph{SIPS: Rebalancing Scale and Angle.}
Cosine-style scoring like X-MoE~\citep{chi2022on} stabilizes routing by removing magnitude factors, but it also discards a potentially informative signal for adaptive gating.
In contrast, SIPS rebalances \emph{scale} and \emph{angle}: it preserves magnitude information for adaptive scoring, while explicitly bounding its influence so that routing remains stable and controllable.
Concretely, SIPS factorizes each logit into a bounded magnitude term and an angular term:
\begin{equation}
    z_i = \phi(\|\bm{q}\|)\,\psi(\|\bm{k}_i\|)\,\cos\theta_i,
    \quad
    \cos\theta_i = \frac{\bm{q}^\top \bm{k}_i}{\|\bm{q}\|\,\|\bm{k}_i\|},
    \label{eq:sips_factored}
\end{equation}
where $\phi(\cdot)$ and $\psi(\cdot)$ are nonnegative scalar transforms. Compared to the previous methods, SIPS bounds the scale sensitivity in both tokens and anchors, while preserving the semantic role of angles and retaining a controlled contribution from magnitudes.

\paragraph{Query Magnitude Transform $\phi(\|\bm{q}\|)$.}
We bound query magnitudes through a \emph{normalize-then-saturate} design.
Normalization is applied first to stabilize the distribution of query norms: for vision models, we apply BatchNorm to the scalar norm $\|\bm{q}\|$, while for LLMs we apply RMSNorm~\citep{zhang2019rmsnorm} to the router input before the projection $\bm{W}_q$.
We then define a bounded query magnitude transform:
\begin{equation}  \phi(\|\bm{q}\|) = \gamma\Big(1+\beta\,\tanh(\widehat{\|\bm{q}\|})\Big),
    \label{eq:phi}
\end{equation}
where $\widehat{\|\bm{q}\|}$ denotes the normalized query norm after the above normalization step.
This yields a controllable interval:
\begin{equation}
    \phi(\|\bm{q}\|)\in[\gamma(1-\beta),\,\gamma(1+\beta)].
\end{equation}
Here, $\beta\in[0,1]$ controls how much magnitude participates in expert selection ($\beta=0$ removes magnitude effects),
and $\gamma$ sets the overall logit scale (equivalently, it rescales logits and modulates the effective softmax temperature).
This design mitigates excessively small logits for low-magnitude queries, while $\tanh(\cdot)$ limits extreme magnitudes and prevents outliers from inducing overly sharp routing, hence improving routing stability.
We provide an additional 3D score-landscape comparison across dot-product, cosine, and SIPS with varying $\beta$ in Appendix~\ref{sec:appendix_phi_beta}, which further illustrates the near-origin smoothness and controlled radial growth induced by query-magnitude saturation.

\paragraph{Anchor Magnitude Transform $\psi(\|\bm{k}_i\|)$.}
Anchor norms regulate inter-expert contribution and competitiveness under top-$k$ aggregation and hence should not be fully removed.
However, to prevent norm-dominated competition, we compress anchor magnitudes around $1$:
\begin{equation}
    \psi(\|\bm{k}_i\|) = 1 + \frac{\|\bm{k}_i\|-1}{p}.
    \label{eq:psi}
\end{equation}
With larger $p$, the marginal logit gain from increasing $\|\bm{k}_i\|$ is dampened, encouraging experts to specialize primarily via angular alignment.
We initialize anchors on the unit sphere ($\|\bm{k}_i\|=1$), so $\psi(\|\bm{k}_i\|)$ is centered at $1$ at initialization and deviations reflect learned competitiveness.

\paragraph{Lipschitz-Controlled Score Geometry.}
A key effect of SIPS is to \emph{explicitly upper-bound the logit sensitivity} to scale variations, yielding a smoother and more controllable scoring geometry.
Recall that in Equation~\eqref{eq:sips_factored},
$z_i = \phi(\|\bm{q}\|)\,\psi(\|\bm{k}_i\|)\,\cos\theta_i$,
the angular term satisfies $\cos\theta_i \in [-1,1]$ and is therefore intrinsically bounded.
Consequently, scale-induced instability in dot-product scoring primarily arises from the magnitude factors $\phi(\|\bm{q}\|)$ and $\psi(\|\bm{k}_i\|)$.
SIPS addresses this by (i) bounding the query scale via $\phi(\|\bm{q}\|)$ (Equation~\eqref{eq:phi}) and (ii) compressing anchor norms around $1$ via $\psi(\|\bm{k}_i\|)$ (Equation~\eqref{eq:psi}).
These designs not only bound the attainable logit magnitude, but also bound the \emph{slope} of $z_i$ with respect to $\|\bm{q}\|$ and $\|\bm{k}_i\|$, which provides explicit control over the Lipschitz behavior of the scoring map along the radial directions.
Geometrically, this reshapes the fixed-anchor score field from an unbounded linear plane (Figure~\ref{fig:sips_heatmap_dot}) to a direction-selective landscape with controlled radial growth (Figure~\ref{fig:sips_heatmap_sips}), mitigating norm-driven ambiguity and improving the stability of expert selection.
Formal gradient-sensitivity analysis and the resulting Lipschitz bounds for Equation~\eqref{eq:sips_factored} are provided in Appendix~\ref{sec:appendix_lipschitz}.

\subsection{Multi-Anchor Heads}
\label{sec:multihead}
While low-rank routing with SIPS provides a stable and discriminative scoring geometry, representing each expert with a single anchor can be restrictive when an expert needs to accommodate multiple semantic modes in the routing space.
To increase routing expressiveness without introducing heavy routing overhead, we propose \textbf{multi-anchor heads}, which associate each expert with multiple anchors in the low-rank routing space.

\paragraph{Multi-Anchor Parameterization.}
For each expert $E_i$, we maintain $H$ anchors $\{\bm{k}_{i,h}\}_{h=1}^{H}$, where $\bm{k}_{i,h}\in\mathbb{R}^{r}$.
Given a routing query $\bm{q}$, we compute anchor-level logits under the SIPS geometry:
\begin{equation}
z_{i,h}
=\phi(\|\bm{q}\|)\,\psi(\|\bm{k}_{i,h}\|)\,\cos\theta_{i,h},
\label{eq:multi_anchor_logit}
\end{equation}
where $\cos\theta_{i,h}
=\frac{\bm{q}^\top \bm{k}_{i,h}}{\|\bm{q}\|\,\|\bm{k}_{i,h}\|}$.
We then aggregate $\{z_{i,h}\}_{h=1}^{H}$ into an expert-level logit $z_i$ via log-sum-exp (LSE) pooling~\citep{jia2012beyond}:
\begin{equation}
    z_i
    =
    \mathrm{LSE}(\{z_{i,h}\}_{h=1}^{H})
    =
    \log\sum_{h=1}^{H}\exp(z_{i,h}),
    \label{eq:multi_anchor_pool}
\end{equation}

Compared to hard $\max$ pooling, LSE provides a smooth approximation that yields more stable gradients while still emphasizing the best-matching anchor.
Routing then proceeds as in Section~\ref{sec:preliminaries} by applying softmax and top-$k$ selection on $\{z_i\}_{i=1}^{N}$.
Geometrically, multi-anchor heads allow each expert to represent a {union of directional regions} in the routing space, improving coverage of multi-view semantics.

\paragraph{Parameter Efficiency.}
Multi-anchor heads only add low-dimensional anchors in $\mathbb{R}^r$ and do not replicate the shared projection $\bm{W}_q$.
As a result, the added parameters scale as $O(NHr)$ and the extra computation is limited to low-rank inner products.
A comprehensive efficiency analysis of L2R is provided in Appendix~\ref{sec:efficiency}.

\paragraph{Initialization.}
To align with the SIPS magnitude transforms, we initialize anchors on the unit sphere in $\mathbb{R}^{r}$ (i.e., $\|\bm{k}_{i,h}\|=1$) so that the initial competition is primarily driven by angular alignment, while norm-based competitiveness is learned gradually through $\psi(\cdot)$.

\begin{figure*}[t]
    \centering

    \begin{subfigure}[t]{0.19\textwidth}
        \centering
        \includegraphics[width=\linewidth]{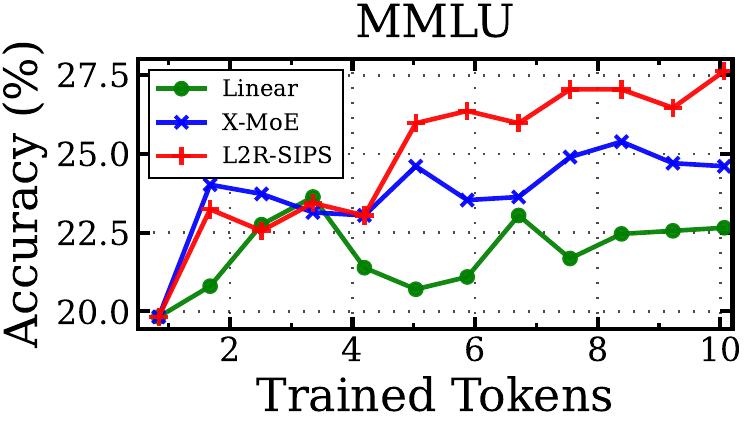}

        \label{fig:panel_mmlu}
    \end{subfigure}
    \hfill
    \begin{subfigure}[t]{0.19\textwidth}
        \centering
        \includegraphics[width=\linewidth]{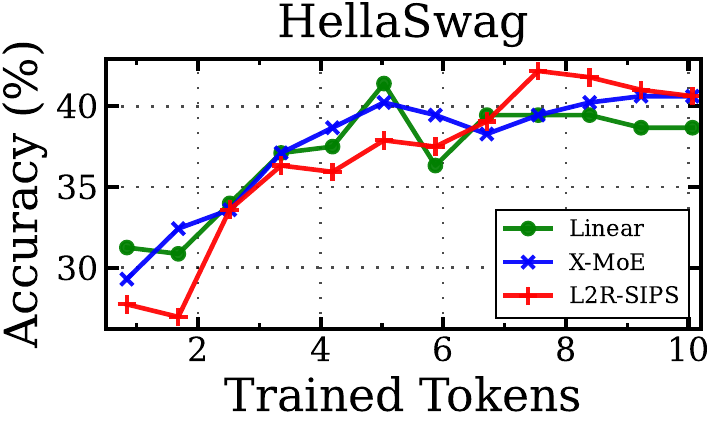}
        \label{fig:panel_hella}
    \end{subfigure}
    \hfill
    \begin{subfigure}[t]{0.19\textwidth}
        \centering
        \includegraphics[width=\linewidth]{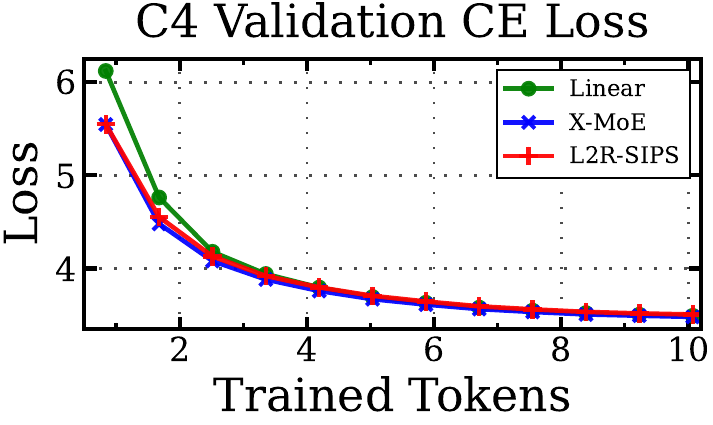}
        \label{fig:panel_c4}
    \end{subfigure}
    \hfill
    \begin{subfigure}[t]{0.19\textwidth}
        \centering
        \includegraphics[width=\linewidth]{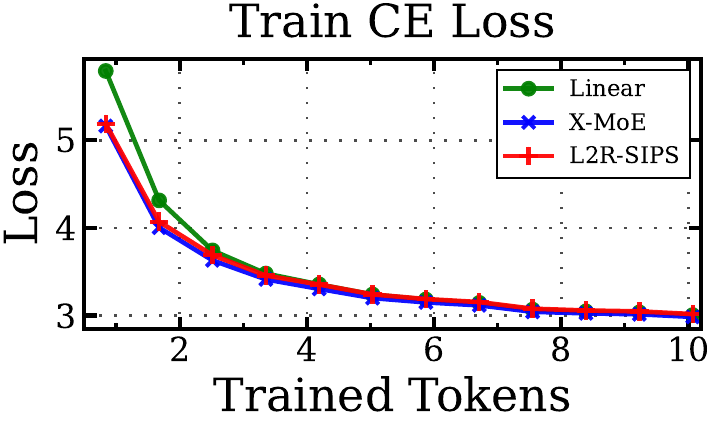}
        \label{fig:panel_traince}
    \end{subfigure}
    \hfill
    \begin{subfigure}[t]{0.19\textwidth}
        \centering
        \includegraphics[width=\linewidth]{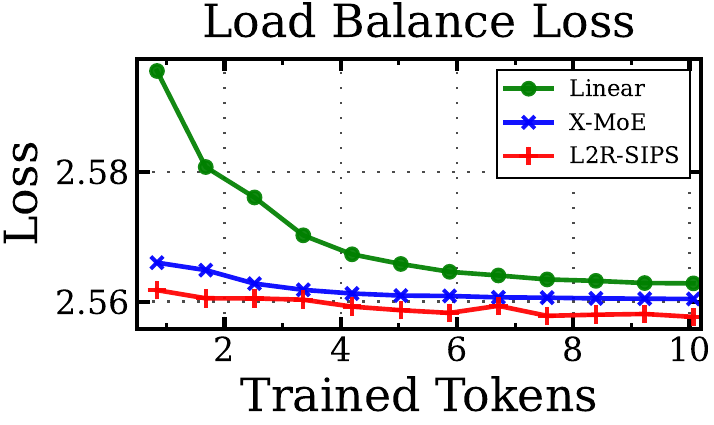}
        \label{fig:panel_lb}
    \end{subfigure}
    
\vspace{-3mm}
    \caption{\textbf{OLMoE training dynamics.}
    Curves show MMLU and HellaSwag accuracies, C4~\citep{raffel2020exploring} validation cross-entropy (CE), training CE, and load-balance loss.
    L2R exhibits clear convergence over 10B tokens, consistently improves
MMLU, and remains competitive on HellaSwag.
    }
    \label{fig:token_scaling_panels}
\vspace{-5mm}
\end{figure*}

\subsection{Optimization Objective}
\label{sec:objective}

L2R is trained under the standard objective used in sparse MoE models, without introducing additional loss terms.
The objective combines the task loss with auxiliary regularizers for routing stability and expert utilization:
\begin{equation}
    \mathcal{L}
    = \mathcal{L}_{\text{task}}
    + \lambda_{\text{bal}}\,\mathcal{L}_{\text{bal}}
    + \lambda_{z}\,\mathcal{L}_{z},
\end{equation}
where $\mathcal{L}_{\text{task}}$ is the task-specific training loss, $\mathcal{L}_{\text{bal}}$ is the standard load-balancing loss~\citep{fedus2021switch} that encourages uniform expert usage, and $\mathcal{L}_{z}$ is the router z-loss~\citep{zoph2022stmoe} for stabilizing routing logits during large-scale LLM training.
The coefficients $\lambda_{\text{bal}}$ and $\lambda_{z}$ weight the auxiliary terms.
Following {OLMoE}~\citep{muennighoff2025olmoe}, we apply $\mathcal{L}_{z}$ for LLM experiments and omit it for vision experiments.
Formal definitions of $\mathcal{L}_{\text{bal}}$ and $\mathcal{L}_{z}$ are provided in Appendix~\ref{sec:objective_details}.

\section{Experiments}
\label{sec:experiments}
This section evaluates L2R on both language and vision MoE systems.
We first study OLMoE-based language MoE training~\citep{muennighoff2025olmoe} and report downstream benchmark performance under the {OLMoE} evaluation protocol.
We then further evaluate L2R on a vision MoE system by training a ViT-S~\citep{deit} backbone on ImageNet-1K~\citep{russakovsky2015imagenet}.

\subsection{LLM Experiments: {OLMoE} Pretraining}
\label{sec:exp_olmoe}

\label{sec:exp_olmoe_setup}
\method{} is evaluated on {OLMoE}~\citep{muennighoff2025olmoe}, an open MoE language model suite.
The backbone and training recipe follow the {OLMoE} reference configuration with $N{=}64$ experts and top-$k{=}8$ sparse activation.
All models are trained \emph{from scratch} for 10B tokens on a distribution-matched subset of the official {OLMoE} pretraining mixture; subset construction and domain statistics are reported in Appendix~\ref{app:olmoe_subset}.
Unless stated otherwise, \method{} uses rank $r{=}2$ and $H{=}16$ multi-anchor heads.
For SIPS, we use $\gamma{=}1$, $\beta{=}1$ for the query transform (Equation~\ref{eq:phi}) and $p{=}4$ for the anchor transform (Equation~\ref{eq:psi}).
Full hyperparameters and implementation details are provided in Appendix~\ref{sec:appendix_olmoe_config}.
All runs are conducted on an $8\times$ NVIDIA B200 setup.

\paragraph{Objective.}
Training minimizes the standard cross entropy loss together with the auxiliary router regularizers adopted by {OLMoE}, including load-balancing loss and router z-loss (Section~\ref{sec:objective}).
Following {OLMoE}, the corresponding weights are set to $\lambda_{\mathrm{bal}}{=}0.01$ and $\lambda_{z}{=}0.001$.

\paragraph{Compared Routers and \method{} Variants.}
The original {OLMoE} linear router is replaced by each of the compared routing designs.
Baselines include the standard {Linear} router and a cosine scoring method {X-MoE}~\citep{chi2022on}.
\method{} is evaluated under different scoring geometries, including {L2R (Dot)}, {L2R (Cosine)}, and the proposed {L2R (SIPS)}.

\paragraph{Downstream Evaluation Protocol.}
We follow the downstream evaluation protocol and reporting conventions of OLMoE~\citep{muennighoff2025olmoe}.
Specifically, we track a fixed set of lightweight commonsense and knowledge benchmarks (e.g., MMLU~\citep{hendrycks2021measuring}, BoolQ~\citep{clark2019boolq}, HellaSwag~\citep{zellers2019hellaswag}, PIQA~\citep{Bisk2019PIQARA}, SciQ~\citep{welbl2017crowdsourcing}) and report {Overall} as the average score across tasks.
This suite is intentionally lightweight for 10B-token training to probe \emph{early-stage} commonsense and knowledge capabilities while remaining stable and reproducible.
Details of task selection and metrics are provided in Appendix~\ref{app:eval_protocol}.

\begin{figure}[t]
    \centering

    \begin{subfigure}[t]{0.30\textwidth}
        \centering
        \includegraphics[width=\linewidth]{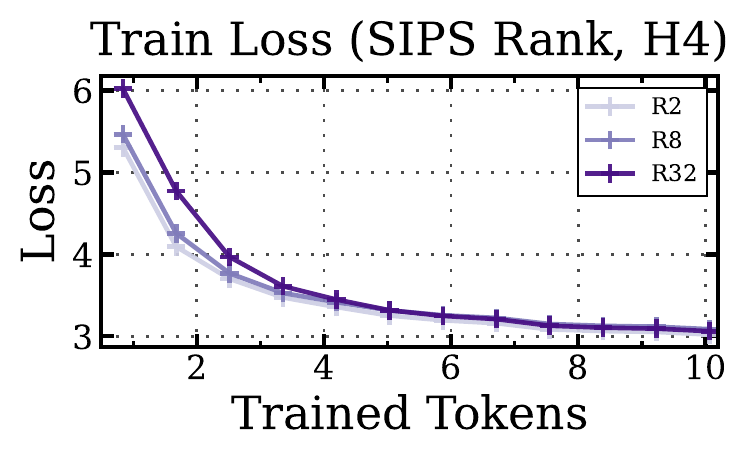}
        \caption{Varying rank with $H{=}4$.}
        \label{fig:ablation_rank_train_loss}
    \end{subfigure}
    \hspace{0.04\textwidth}
    \begin{subfigure}[t]{0.30\textwidth}
        \centering
        \includegraphics[width=\linewidth]{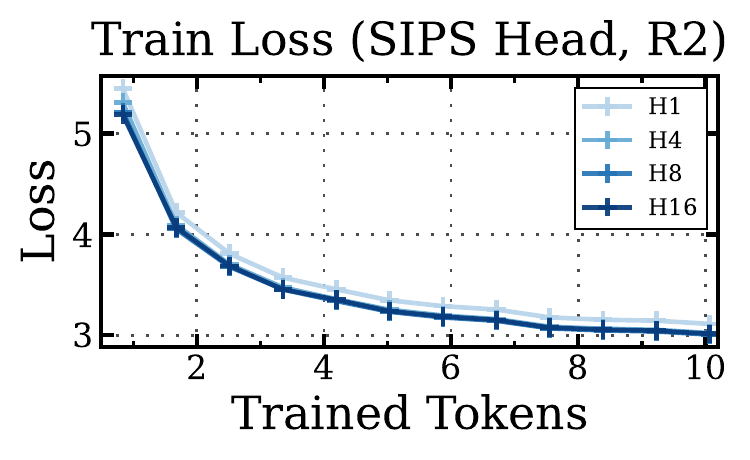}
        \caption{Varying heads with $r{=}2$.}
        \label{fig:ablation_head_train_loss}
    \end{subfigure}
    \vspace{-2mm}
    \caption{\textbf{Training dynamics for ablations.}
    In the evaluated settings, lower rank and more heads are associated with faster training-loss convergence.}
    \label{fig:ablation_train_loss}
    \vspace{-3mm}
\end{figure}

\subsubsection{Main Results}
\label{sec:exp_olmoe_main}

\paragraph{Training Dynamics.}
Figure~\ref{fig:token_scaling_panels} summarizes training behavior over 10B tokens.
All routers show smooth and monotonic loss reduction, confirming stable convergence under the OLMoE recipe.
Throughout training, \method{} consistently attains higher MMLU accuracy and remains competitive on HellaSwag.
For both C4~\citep{raffel2020exploring} validation CE and training CE, \method{} closely tracks the angle-sensitive baseline (X-MoE) and converges faster than the Linear router, supporting the benefit of stronger angular discrimination for optimization and generalization.
Finally, in terms of load-balance loss, \method{} achieves faster and lower convergence, indicating more balanced expert utilization and stable routing dynamics.

\begin{wraptable}{r}{0.6\textwidth}
\centering
\vspace{-1.2em}
\caption{Overall performance.}
\label{tab:main_results}
\vspace{-1mm}
\scriptsize
\setlength{\tabcolsep}{2.5pt}
\renewcommand{\arraystretch}{0.88}
\resizebox{\linewidth}{!}{%
\begin{tabular}{lcccccc}
\toprule
Method & MMLU & BoolQ & HSwag & PIQA & SciQ & Overall \\
\midrule
Linear & 22.7 & 60.0 & 38.7 & 61.1 & 66.8 & 39.6 \\
Linear (SIPS) & 24.2 & \textbf{61.7} & \textbf{42.6} & \underline{63.5} & 68.4 & 41.6 \\
\midrule
X-MoE & 24.6 & 60.0 & 40.6 & \textbf{64.5} & 68.8 & 41.5 \\
L2R (Dot) & 24.4 & 59.2 & 39.5 & 60.5 & 64.8 & 40.3 \\
L2R (Cosine) & \underline{24.9} & \underline{61.1} & \underline{42.2} & 60.0 & \textbf{72.7} & \underline{42.0} \\
\midrule
\textbf{L2R (SIPS)} & \textbf{27.6} & \textbf{61.7} & 40.6 & 62.3 & \underline{71.5} & \textbf{43.3} \\
\bottomrule
\end{tabular}
}

\begin{minipage}{0.96\linewidth}
\scriptsize
\emph{Note:} Best results are in \textbf{bold}, and second-best results are \underline{underlined}.
\end{minipage}
\vspace{-3mm}
\end{wraptable}

\paragraph{Overall Performance.}
Table~\ref{tab:main_results} reports downstream evaluation results at 10B tokens.
To isolate the effect of SIPS, we apply SIPS directly to the full-rank Linear router without low-rank projection or multi-anchor heads. Linear (SIPS) improves the overall score from $39.6$ to $41.6$, showing that SIPS alone can provide a clear gain over dot-product scoring.
Meanwhile, \method{} (Dot) achieves a modest gain over Linear ($40.3$ vs.\ $39.6$) while retaining dot-product scoring, suggesting a benefit from low-rank routing even without SIPS.
Combining low-rank routing, SIPS, and multi-anchor heads, \method{} (SIPS) achieves the best overall score ($43.3$), outperforming both Linear (SIPS) ($41.6$) and X-MoE ($41.5$).
Notably, \method{} (SIPS) yields a substantial gain on MMLU, improving from $22.7$ (Linear) and $24.6$ (X-MoE) to $27.6$, indicating stronger early-stage knowledge recall and reasoning on lightweight probes.

\paragraph{Angle Discrimination vs. Scale Domination.}
The results reveal a consistent split between (i) \emph{angle-sensitive} routing and (ii) \emph{unconstrained dot-product} routing.
Angle-sensitive designs (X-MoE, L2R (Cosine), and \method{} (SIPS)) substantially outperform Linear and L2R (Dot) in overall quality, and also exhibit faster cross-entropy reduction during training (Figure~\ref{fig:token_scaling_panels}), underscoring the importance of angular discrimination for reliable routing.
Meanwhile, pure cosine scoring eliminates magnitude cues, whereas SIPS retains them with bounded influence.
This controlled rebalancing yields stronger overall performance over X-MoE and L2R (Cosine), supporting the role of \emph{controlled} magnitude in expressive yet stable expert aggregation.

\subsubsection{Ablation Study} 
\label{sec:exp_olmoe_ablation}

\begin{wraptable}{r}{0.45\textwidth}
\centering
\vspace{-1.2em}
\caption{Ablation on routing rank $r$ with $H=4$.}
\label{tab:ablation_rank}
\vspace{-1mm}
\small
\setlength{\tabcolsep}{3pt}
\renewcommand{\arraystretch}{0.9}
\resizebox{\linewidth}{!}{%
\begin{tabular}{ccccccc}
\toprule
Rank & MMLU & BoolQ & HSwag & PIQA & SciQ & Overall \\
\midrule
2  & 25.2 & 62.1 & \textbf{39.5} & 61.1 & 69.1 & \underline{41.6} \\
8  & 23.2 & \textbf{62.3} & 38.3 & \textbf{63.1} & 68.8 & 40.7 \\
32 & \textbf{25.6} & 59.4 & 39.1 & 61.1 & \textbf{71.9} & \textbf{41.7} \\
\bottomrule
\end{tabular}
}
\vspace{-5mm}
\end{wraptable}

\paragraph{Ablation on Routing Rank $r$.}
Table~\ref{tab:ablation_rank} compares routing ranks with $H{=}4$.
Although $r{=}32$ achieves a slightly higher overall score, $r{=}2$ delivers comparable downstream performance and exhibits faster training-loss convergence (Figure~\ref{fig:ablation_rank_train_loss}).
In the vision rank ablation (Section~\ref{sec:exp_vision}), $r{=}2$ achieves the best Top-1 and Top-5 accuracy among the evaluated ranks.
Considering these results jointly, we adopt $r{=}2$ as a shared default rank for both language and vision experiments, aiming for a consistent setting that performs well across domains rather than selecting a separate rank for each domain.

\begin{wraptable}{r}{0.66\textwidth}
\centering
\vspace{-1.2em}
\caption{Ablation on multi-anchor heads $H$ with rank $r{=}2$.}
\label{tab:ablation_heads}
\vspace{-1mm}
\scriptsize
\setlength{\tabcolsep}{2.6pt}
\renewcommand{\arraystretch}{0.88}
\resizebox{\linewidth}{!}{%
\begin{tabular}{cccccccc}
\toprule
$H$ & Router Params & MMLU & BoolQ & HSwag & PIQA & SciQ & Overall \\
\midrule
1  & 100.4K (4.79\%) & 25.5 & 60.0 & 37.5 & 60.5 & 69.9 & 41.2 \\
4  & 106.5K (5.08\%) & 25.2 & 62.1 & 39.5 & 61.1 & 69.1 & 41.6 \\
8  & 114.7K (5.47\%) & 25.2 & 60.2 & 39.8 & \textbf{64.3} & 68.4 & 41.7 \\
16 & 131.1K (6.25\%) & \textbf{27.6} & 61.7 & \textbf{40.6} & 62.3 & \textbf{71.5} & \textbf{43.3} \\
64 & 196.6K (9.38\%) & 26.9 & \textbf{62.1} & 40.2 & 62.1 & 68.8 & 42.3 \\
\bottomrule
\end{tabular}
}

\begin{minipage}{0.96\linewidth}
\scriptsize
\emph{Note:} Router parameters are reported over all MoE layers, with percentages relative to the Linear router.
\end{minipage}
\vspace{-2mm}
\end{wraptable}

\paragraph{Ablation on Multi-Anchor Heads $H$.}
Table~\ref{tab:ablation_heads} varies the head count with $r{=}2$.
Increasing $H$ from $1$ to $16$ steadily improves overall performance.
Within this range, the training-loss curves in Figure~\ref{fig:ablation_head_train_loss} also suggest faster convergence with more heads.
However, further increasing $H$ to $64$ does not yield additional downstream gains, with $H{=}16$ achieving the best overall score among the evaluated settings.
This suggests a practical head-selection guideline: a moderate number of heads, such as $H{=}8$ or $H{=}16$, provides a favorable trade-off between multi-view routing capacity and anchor redundancy.
These results support the effectiveness of the multi-anchor routing pattern, which enriches routing flexibility in the low-rank space without incurring prohibitive router overhead.

\subsubsection{Routing Geometry and Expert Usage}
\label{sec:exp_olmoe_viz}

\begin{wrapfigure}{r}{0.7\textwidth}
    \centering
    \vspace{-0.5em}
    \centering
    \begin{subfigure}[t]{0.32\linewidth}
        \centering
        \includegraphics[width=0.91\linewidth]{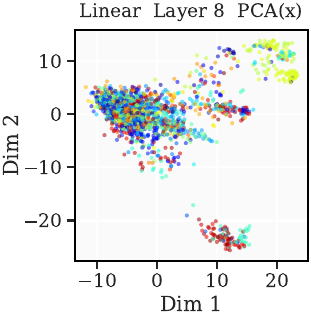}
        \caption{Linear: PCA($\bm{x}$)}
        \label{fig:viz_linear_pca_x}
    \end{subfigure}
    \hfill
    \begin{subfigure}[t]{0.32\linewidth}
        \centering
        \includegraphics[width=\linewidth]{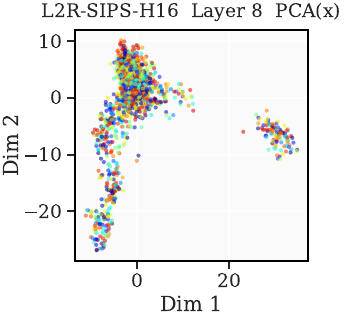}
        \caption{\method{}: PCA($\bm{x}$)}
        \label{fig:viz_l2r_pca_x}
    \end{subfigure}
    \hfill
    \begin{subfigure}[t]{0.32\linewidth}
        \centering
        \includegraphics[width=\linewidth]{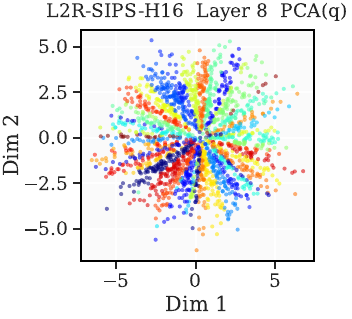}
        \caption{\method{}: PCA($\bm{q}$)}
        \label{fig:viz_l2r_pca_q}
    \end{subfigure}
    \caption{\textbf{Representation-space visualization.}
    PCA projections of token representations in the raw backbone space ($\bm{x}$) and the low-rank routing space ($\bm{q}$).
    Points are colored by the {top-1 routed expert}.
    }
    \label{fig:viz_rep_space}
    \vspace{-2mm}
\end{wrapfigure}

\paragraph{Routing-Space Geometry.}
Figure~\ref{fig:viz_rep_space} contrasts token geometry in the raw backbone space $\bm{x}$ and the projected routing space $\bm{q}$.
Under Linear routing (Figure~\ref{fig:viz_linear_pca_x}), tokens concentrate in an anisotropic and ``narrow'' region, consistent with the severe angular concentration quantified in Figure~\ref{fig:concentration}.
In contrast, \method{} learns a routing-oriented low-rank space where token directions are markedly more separable and form structured regions when stratified by top-1 expert assignments (Figure~\ref{fig:viz_l2r_pca_q}).
Tracing these same tokens back to the raw $\bm{x}$ space reveals wide dispersion  (Figure~\ref{fig:viz_l2r_pca_x}), which is consistent with a multi-anchor specialization pattern that each expert covers multiple semantic modes instead of a single compact cluster in $\bm{x}$.

\paragraph{Near-Origin Stability with SIPS.}
Both the raw and projected routing spaces contain low-magnitude tokens around the origin, where routing is inherently noise-sensitive (cosine-style scoring can be particularly brittle to small angular perturbations).
SIPS stabilizes this regime by imposing a bounded and smooth magnitude response, avoiding vanishing scores for low-norm queries, thereby improving stability without discarding magnitude information.
Further diagnostics and ablations on $\phi(\cdot)$ are provided in Appendix~\ref{sec:appendix_phi_beta}.

\begin{figure}[htbp]
    \centering
    \begin{subfigure}[t]{0.31\linewidth}
        \includegraphics[width=\linewidth]{figures/usage/ours__load_heatmap__top1__plotly.pdf}
    \end{subfigure}
    \hfill
    \begin{subfigure}[t]{0.31\linewidth}
        \includegraphics[width=\linewidth]{figures/usage/ours__load_heatmap__topk__plotly.pdf}
    \end{subfigure}
    \hfill
    \begin{subfigure}[t]{0.31\linewidth}
        \includegraphics[width=\linewidth]{figures/usage/ours__load_heatmap__importance__plotly.pdf}
    \end{subfigure}
    
    \begin{subfigure}[t]{0.31\linewidth}
        \includegraphics[width=\linewidth]{figures/usage/linear__load_heatmap__top1__plotly.pdf}
    \end{subfigure}
    \hfill
    \begin{subfigure}[t]{0.31\linewidth}
        \includegraphics[width=\linewidth]{figures/usage/linear__load_heatmap__topk__plotly.pdf}
    \end{subfigure}
    \hfill
    \begin{subfigure}[t]{0.31\linewidth}
        \includegraphics[width=\linewidth]{figures/usage/linear__load_heatmap__importance__plotly.pdf}
    \end{subfigure}
    \caption{\textbf{Expert-usage patterns.} Each row is a Transformer layer, and each column is an expert.
    From left to right: top-1 frequency, top-$k$ frequency, and importance-based routing weights.
    }
    \label{fig:viz_expert_usage}
    \vspace{-4mm}
\end{figure}

\paragraph{Expert Usage and Cooperation.}
Figure~\ref{fig:viz_expert_usage} summarizes routing patterns across layers.
Compared to Linear routing, \method{} yields a sharper and more structured top-1 map, indicating clearer expert specialization, while maintaining broadly distributed top-$k$ usage, reflecting sustained cooperation under sparse activation.
Importantly, regions with high top-1 frequency do not consistently align with peaks in the top-$k$ and importance maps, suggesting that \method{} avoids single-expert domination and instead modulates multi-expert contributions via SIPS-controlled magnitude. 

\begin{wraptable}{r}{0.55\textwidth}
\centering
\caption{Quantitative routing diagnostics on OLMoE.}
\label{tab:routing_diagnostics_main}
\small
\setlength{\tabcolsep}{3pt}
\renewcommand{\arraystretch}{1.05}
\begin{tabular}{lcccc}
\toprule
Model
& \makecell{Mean\\Margin $\uparrow$}
& \makecell{Low-Margin\\Rate $\downarrow$}
& \makecell{Stability\\$\uparrow$}
& \makecell{Top-$k$\\Overlap $\uparrow$} \\
\midrule
X-MoE
& 0.0712 & 0.9424 & 0.9566 & 0.9683 \\
L2R (Dot)-H1
& 0.0105 & 0.9989 & 0.6734 & 0.8311 \\
L2R (SIPS)-H1
& \textbf{0.1806} & \textbf{0.7036}
& \textbf{0.9847} & \textbf{0.9915} \\
\bottomrule
\end{tabular}
\end{wraptable}

\paragraph{Quantitative Routing Diagnostics.}
Table~\ref{tab:routing_diagnostics_main} complements the visualizations
with expert-discrimination and routing-robustness metrics.
Mean Margin is the average top-1--top-2 logit gap, and Low-Margin Rate
is the fraction of tokens with a gap below $0.2$.
Under Gaussian noise with $\sigma=0.02$ applied to router inputs,
Stability measures the fraction of unchanged top-1 assignments,
while Top-$k$ Overlap measures the Jaccard similarity of the
selected top-8 expert sets.
For the single-anchor variants, replacing dot-product scoring
with SIPS increases the mean margin from $0.0105$ to $0.1806$
and stability from $0.6734$ to $0.9847$.
These results support improved expert discrimination and routing
robustness in the evaluated setting.
Full results and metric definitions are provided in
Appendix~\ref{app:routing_diagnostics}.

\subsection{Experiments with ViT on ImageNet}
\label{sec:exp_vision}

\begin{wraptable}{r}{0.4\textwidth}
\centering
\vspace{-1.2em}
\caption{Top-1 and Top-5 accuracy on ImageNet-1K with ViT-S backbone.}\vspace{-1mm}
\label{tab:imagenet-vit-s}
\centering
\resizebox{\linewidth}{!}{
\centering
\setlength{\tabcolsep}{6pt}
\renewcommand{\arraystretch}{0.9}
\begin{tabular}{lcc}
\toprule
{Method} & {Top-1 (\%)} $\uparrow$ & {Top-5 (\%)} $\uparrow$ \\
\midrule
Linear & 74.12 & 90.38 \\
X-MoE & 74.07 & 90.23 \\
\midrule
L2R (Dot) & 74.96 & 91.23 \\
L2R (Cosine) & 75.32 & 91.36 \\
\midrule
\textbf{L2R (SIPS)} & \textbf{76.58} & \textbf{91.50} \\
\bottomrule
\end{tabular}}
\vspace{-2mm}
\end{wraptable}

We further evaluate \method{} on a vision MoE system by training a ViT-S~\citep{deit} backbone on ImageNet-1K~\citep{russakovsky2015imagenet} \emph{from scratch} for 300 epochs.
We follow a standard ViT-S recipe with batch size 512 and apply MoE to the last 4 Transformer blocks.
Unless stated otherwise, we use $N{=}16$ experts with top-$k{=}2$ sparse activation, and train with only the load-balancing auxiliary loss ($\lambda_{\mathrm{bal}}{=}0.01$).
For \method{}, we set rank $r{=}2$ and multi-anchor heads $H{=}4$, and reuse the same SIPS hyperparameters as in the LLM setting ($\gamma{=}1$, $\beta{=}1$, $p{=}4$).
All runs are conducted on $4\times$ A100 GPUs. Detailed configurations are summarized in Appendix~\ref{sec:appendix_vit_config}.

\paragraph{Performance on ImageNet.}
Table~\ref{tab:imagenet-vit-s} reports ImageNet results.
\method{} achieves the best performance, suggesting that L2R with the proposed SIPS transfers effectively to vision MoE training.
Notably, X-MoE~\citep{chi2022on} slightly underperforms the Linear router in this setting, whereas L2R variants yield consistent gains, suggesting that the proposed routing design is beneficial beyond the language domain.

\begin{wraptable}{r}{0.50\textwidth}
\centering
\vspace{-1.2em}
\caption{Ablations on routing rank $r$ and multi-anchor heads $H$ on ImageNet-1K.}
\label{tab:imagenet_ablation}
\vspace{-1mm}
\small
\setlength{\tabcolsep}{3.5pt}
\renewcommand{\arraystretch}{0.9}
\resizebox{\linewidth}{!}{%
\begin{tabular}{ccc@{\hspace{8pt}}ccc}
\toprule
\multicolumn{3}{c}{Routing Rank ($H{=}4$)}
&
\multicolumn{3}{c}{Multi-Anchor Heads ($r{=}2$)}
\\
\cmidrule(lr){1-3}\cmidrule(lr){4-6}
$r$ & Top-1 (\%) $\uparrow$ & Top-5 (\%) $\uparrow$
&
$H$ & Top-1 (\%) $\uparrow$ & Top-5 (\%) $\uparrow$
\\
\midrule
2  & \textbf{76.58} & \textbf{91.50}
&
1  & 75.47 & 91.31
\\
8  & 75.44 & 90.95
&
4  & \textbf{76.58} & 91.50
\\
32 & 74.90 & 90.12
&
8  & 76.22 & \textbf{91.79}
\\
-- & -- & --
&
16 & 75.20 & 91.00
\\
\bottomrule
\end{tabular}%
}
\vspace{-5mm}
\end{wraptable}

\paragraph{Ablations on Routing Rank and Multi-Anchor Heads.}
Table~\ref{tab:imagenet_ablation} examines the two main architectural
choices of \method{} in the vision setting.
With $H{=}4$, $r{=}2$ achieves the best Top-1 and Top-5 accuracy,
while increasing the routing rank to $8$ or $32$ reduces both metrics.
Together with the comparable downstream performance of $r{=}2$ and $r{=}32$ in the LLM rank ablation (Table~\ref{tab:ablation_rank}), these vision results support adopting $r{=}2$ as a shared default rank across language and vision, rather than selecting a separate rank for each domain.
With $r{=}2$, $H{=}4$ gives the highest Top-1 accuracy, while $H{=}8$
gives the highest Top-5 accuracy; increasing $H$ to $16$ reduces both.
Together, these results support a compact routing space with moderate
anchor multiplicity in the evaluated vision setting.

\section{Conclusion}
\label{sec:conclusion}
We introduced \textbf{L}ow-rank \& \textbf{L}ipschitz-controlled \textbf{R}outing (\textbf{L2R}), a routing framework that improves MoE specialization by jointly redesigning the routing space and scoring geometry. L2R replaces high-dimensional linear routing with a shared low-rank latent space, adopts Saturated Inner-Product Scoring (\textbf{SIPS}) to bound magnitude sensitivity while preserving controlled magnitude for adaptive expert weighting, and extends expressiveness with lightweight multi-anchor heads. Experiments on OLMoE pretraining and ViT-based ImageNet training show
consistent gains on most benchmarks, with OLMoE results further showing
faster convergence and balanced expert usage, indicating that L2R provides
a practical routing recipe across both language and vision MoE systems.

\bibliographystyle{tmlr}
\bibliography{main}

@inproceedings{muennighoff2025olmoe,
  title     = {{OLMoE}: Open Mixture-of-Experts Language Models},
  author={Muennighoff, Niklas and Soldaini, Luca and Groeneveld, Dirk and Lo, Kyle and Morrison, Jacob and Min, Sewon and Shi, Weijia and Walsh, Pete and Tafjord, Oyvind and Lambert, Nathan and others},
  booktitle = {Proc. ICLR},
  year      = {2025}
}

@article{deepseekmoe,
  title={Deepseekmoe: Towards ultimate expert specialization in mixture-of-experts language models},
  author={Dai, Damai and Deng, Chengqi and Zhao, Chenggang and Xu, RX and Gao, Huazuo and Chen, Deli and Li, Jiashi and Zeng, Wangding and Yu, Xingkai and Wu, Yu and others},
  journal={arXiv preprint arXiv:2401.06066},
  year={2024}
}

@inproceedings{vmoe,
  title={Scaling vision with sparse mixture of experts},
  author={Riquelme, Carlos and Puigcerver, Joan and Mustafa, Basil and Neumann, Maxim and Jenatton, Rodolphe and Susano Pinto, Andr{\'e} and Keysers, Daniel and Houlsby, Neil},
  booktitle = {Proc. NeurIPS},
  pages={8583--8595},
  year={2021}
}

@inproceedings{chi2022on,
  title={On the representation collapse of sparse mixture of experts},
  author={Chi, Zewen and Dong, Li and Huang, Shaohan and Dai, Damai and Ma, Shuming and Patra, Barun and Singhal, Saksham and Bajaj, Payal and Song, Xia and Mao, Xian-Ling and others},
  booktitle = {Proc. NeurIPS},
  pages={34600--34613},
  year= {2022}
}

@inproceedings{dikkala2023routing,
  title={On the benefits of learning to route in mixture-of-experts models},
  author={Dikkala, Nishanth and Ghosh, Nikhil and Meka, Raghu and Panigrahy, Rina and Vyas, Nikhil and Wang, Xin},
  booktitle = {Proc. EMNLP},
  pages={9376--9396},
  year={2023}
}

@article{jiang2024mixtral,
  title={Mixtral of experts},
  author={Jiang, Albert Q and Sablayrolles, Alexandre and Roux, Antoine and Mensch, Arthur and Savary, Blanche and Bamford, Chris and Chaplot, Devendra Singh and Casas, Diego de las and Hanna, Emma Bou and Bressand, Florian and others},
  journal={arXiv preprint arXiv:2401.04088},
  year={2024}
}

@inproceedings{chen2023modsquad,
  title={Mod-squad: Designing mixtures of experts as modular multi-task learners},
  author={Chen, Zitian and Shen, Yikang and Ding, Mingyu and Chen, Zhenfang and Zhao, Hengshuang and Learned-Miller, Erik G and Gan, Chuang},
  booktitle = {Proc. CVPR},
  pages={11828--11837},
  year={2023}
}

@article{yang2025ase,
  title   = {Adaptive Shared Experts with {LoRA}-Based Mixture of Experts for Multi-Task Learning},
  author  = {Yang, Minghao and Togo, Ren and Li, Guang and Ogawa, Takahiro and Haseyama, Miki},
  journal = {arXiv preprint arXiv:2510.00570},
  year    = {2025}
}

@article{cai2013angle,
  title={Distributions of angles in random packing on spheres},
  author={Cai, Tony and Fan, Jianqing and Jiang, Tiefeng},
  journal = {J. Mach. Learn. Res.},
  volume={14},
  number={1},
  pages={1837--1864},
  year={2013},
}

@article{cai2025survey,
  title={A survey on mixture of experts in large language models},
  author={Cai, Weilin and Jiang, Juyong and Wang, Fan and Tang, Jing and Kim, Sunghun and Huang, Jiayi},
  journal = {IEEE Trans. Knowl. Data Eng.},
  year    = {2025}
}

@article{yang2024qwen2,
  title   = {{Qwen2} Technical Report},
  author  = {Yang, An and Yang, Baosong and Hui, Binyuan and others},
  journal = {arXiv preprint arXiv:2407.10671},
  year    = {2024}
}

@article{wu2024yuan20m32mixtureexperts,
  title={Yuan 2.0-m32: Mixture of experts with attention router},
  author={Wu, Shaohua and Luo, Jiangang and Chen, Xi and Li, Lingjun and Zhao, Xudong and Yu, Tong and Wang, Chao and Wang, Yue and Wang, Fei and Qiao, Weixu and others},
  journal={arXiv preprint arXiv:2405.17976},
  year={2024}
}

@article{ran2025routerupcycling,
  title={Router Upcycling: Leveraging Mixture-of-Routers in Mixture-of-Experts Upcycling},
  author={Ran, Junfeng and Zhao, Guangxiang and Wu, Yuhan and Zhu, Dawei and Wu, Longyun and Zhao, Yikai and Yang, Tong and Sun, Lin and Zhang, Xiangzheng and Li, Sujian},
  journal={arXiv preprint arXiv:2509.00679},
  year={2025}
}

@article{elhage2022toy,
  title={Toy models of superposition},
  author={Elhage, Nelson and Hume, Tristan and Olsson, Catherine and Schiefer, Nicholas and Henighan, Tom and Kravec, Shauna and Hatfield-Dodds, Zac and Lasenby, Robert and Drain, Dawn and Chen, Carol and others},
  journal={arXiv preprint arXiv:2209.10652},
  year={2022}
}

@inproceedings{nguyen2024statistical,
  title={Statistical advantages of perturbing cosine router in mixture of experts},
  author={Nguyen, Huy and Akbarian, Pedram and Pham, Trang and Nguyen, Trang and Zhang, Shujian and Ho, Nhat},
  booktitle = {Proc. ICLR},
  year      = {2025}
}

@inproceedings{
wu2024multihead,
  title={Multi-head mixture-of-experts},
  author={Wu, Xun and Huang, Shaohan and Wang, Wenhui and Ma, Shuming and Dong, Li and Wei, Furu},
booktitle = {Proc. NeurIPS},
  pages={94073--94096},
  year={2024}
}

@inproceedings{zhang2019rmsnorm,
  title={Root Mean Square Layer Normalization},
  author={Zhang, Biao and Sennrich, Rico},
booktitle = {Proc. NeurIPS},
  year={2019}
}

@inproceedings{gu2025olmes,
    title = "{OLMES}: A Standard for Language Model Evaluations",
    author = "Gu, Yuling  and
      Tafjord, Oyvind  and
      Kuehl, Bailey  and
      Haddad, Dany  and
      Dodge, Jesse  and
      Hajishirzi, Hannaneh",
  booktitle = {Proc. NAACL},
  pages = "5005--5033",
  year      = {2025}
}

@article{clark2018think,
  title={Think you have solved question answering? try arc, the ai2 reasoning challenge},
  author={Clark, Peter and Cowhey, Isaac and Etzioni, Oren and Khot, Tushar and Sabharwal, Ashish and Schoenick, Carissa and Tafjord, Oyvind},
  journal={arXiv preprint arXiv:1803.05457},
  year={2018}
}

@article{clark2019boolq,
  title={Boolq: Exploring the surprising difficulty of natural yes/no questions},
  author={Clark, Christopher and Lee, Kenton and Chang, Ming-Wei and Kwiatkowski, Tom and Collins, Michael and Toutanova, Kristina},
  journal={arXiv preprint arXiv:1905.10044},
  year={2019}
}

@article{zellers2019hellaswag,
  title={Hellaswag: Can a machine really finish your sentence?},
  author={Zellers, Rowan and Holtzman, Ari and Bisk, Yonatan and Farhadi, Ali and Choi, Yejin},
  journal={arXiv preprint arXiv:1905.07830},
  year={2019}
}

@inproceedings{hendrycks2021measuring,
  title={Measuring massive multitask language understanding},
  author={Hendrycks, Dan and Burns, Collin and Basart, Steven and Zou, Andy and Mazeika, Mantas and Song, Dawn and Steinhardt, Jacob},
  booktitle = {Proc. ICLR},
  year      = {2021}
}

@inproceedings{Bisk2019PIQARA,
  title={Piqa: Reasoning about physical commonsense in natural language},
  author={Bisk, Yonatan and Zellers, Rowan and Gao, Jianfeng and Choi, Yejin and others},
  booktitle = {Proc. AAAI},
  volume={34},
  pages={7432--7439},
  year={2020}
}

@article{welbl2017crowdsourcing,
  title={Crowdsourcing multiple choice science questions},
  author={Welbl, Johannes and Liu, Nelson F and Gardner, Matt},
  journal={arXiv preprint arXiv:1707.06209},
  year={2017}
}

@inproceedings{sakaguchi2019winogrande,
  title={Winogrande: An adversarial winograd schema challenge at scale},
  author={Sakaguchi, Keisuke and Le Bras, Ronan and Bhagavatula, Chandra and Choi, Yejin},
  booktitle = {Proc. AAAI},
  volume={34},
  pages={8732--8740},
  year={2020}
}

@inproceedings{gale2022megablocks,
  title={Megablocks: Efficient sparse training with mixture-of-experts},
  author={Gale, Trevor and Narayanan, Deepak and Young, Cliff and Zaharia, Matei},
  booktitle = {Proc. MLSys},
  volume={5},
  pages={288--304},
  year={2023}
}

@inproceedings{hwang2023tutel,
  title={Tutel: Adaptive mixture-of-experts at scale},
  author={Hwang, Changho and Cui, Wei and Xiong, Yifan and Yang, Ziyue and Liu, Ze and Hu, Han and Wang, Zilong and Salas, Rafael and Jose, Jithin and Ram, Prabhat and others},
  booktitle = {Proc. MLSys},
  volume={5},
  pages={269--287},
  year={2023},
}

@inproceedings{jia2012beyond,
  title={Beyond spatial pyramids: Receptive field learning for pooled image features},
  author={Jia, Yangqing and Huang, Chang and Darrell, Trevor},
  booktitle = {Proc. CVPR},
  pages={3370--3377},
  year      = {2012}
}

@article{zoph2022stmoe,
  title={St-moe: Designing stable and transferable sparse expert models},
  author={Zoph, Barret and Bello, Irwan and Kumar, Sameer and Du, Nan and Huang, Yanping and Dean, Jeff and Shazeer, Noam and Fedus, William},
  journal={arXiv preprint arXiv:2202.08906},
  year={2022}
}

@article{russakovsky2015imagenet,
  title={Imagenet large scale visual recognition challenge},
  author={Russakovsky, Olga and Deng, Jia and Su, Hao and Krause, Jonathan and Satheesh, Sanjeev and Ma, Sean and Huang, Zhiheng and Karpathy, Andrej and Khosla, Aditya and Bernstein, Michael and others},
  journal = {Int. J. Comput. Vis.},
  volume={115},
  number={3},
  pages={211--252},
  year={2015},
}

@inproceedings{deit,
  title={Training data-efficient image transformers \& distillation through attention},
  author={Touvron, Hugo and Cord, Matthieu and Douze, Matthijs and Massa, Francisco and Sablayrolles, Alexandre and J{\'e}gou, Herv{\'e}},
  booktitle = {Proc. ICML},
  pages =     {10347--10357},
  year =      {2021},
}

@article{shazeer2017outrageously,
  title={Outrageously large neural networks: The sparsely-gated mixture-of-experts layer},
  author={Shazeer, Noam and Mirhoseini, Azalia and Maziarz, Krzysztof and Davis, Andy and Le, Quoc and Hinton, Geoffrey and Dean, Jeff},
  journal={arXiv preprint arXiv:1701.06538},
  year={2017}
}

@article{lepikhin2020gshard,
  title={Gshard: Scaling giant models with conditional computation and automatic sharding},
  author={Lepikhin, Dmitry and Lee, HyoukJoong and Xu, Yuanzhong and Chen, Dehao and Firat, Orhan and Huang, Yanping and Krikun, Maxim and Shazeer, Noam and Chen, Zhifeng},
  journal={arXiv preprint arXiv:2006.16668},
  year={2020}
}

@article{fedus2021switch,
  title={Switch transformers: Scaling to trillion parameter models with simple and efficient sparsity},
  author={Fedus, William and Zoph, Barret and Shazeer, Noam},
  journal={J. Mach. Learn. Res.},
  volume={23},
  number={120},
  pages={1--39},
  year={2022}
}

@article{raffel2020exploring,
  title={Exploring the limits of transfer learning with a unified text-to-text transformer},
  author={Raffel, Colin and Shazeer, Noam and Roberts, Adam and Lee, Katherine and Narang, Sharan and Matena, Michael and Zhou, Yanqi and Li, Wei and Liu, Peter J},
  journal = {J. Mach. Learn. Res.},
  volume={21},
  number={140},
  pages={1--67},
  year={2020}
}

\newpage
\appendix

\section{Quantitative Diagnostics of Routing Behavior}
\label{app:routing_diagnostics}

\subsection{Quantitative Metrics for Expert Discrimination}
\label{app:expert_discrimination}
\begin{wraptable}{r}{0.48\textwidth}
\centering
\vspace{-1.2em}
\caption{Direct expert-discrimination metrics.}
\label{tab:expert_discrimination}
\vspace{-1mm}
\small
\setlength{\tabcolsep}{4pt}
\renewcommand{\arraystretch}{0.95}
\resizebox{\linewidth}{!}{%
\begin{tabular}{lcc}
\toprule
Model & Mean Margin $\uparrow$ & Low-Margin Rate $\downarrow$ \\
\midrule
X-MoE & 0.0712 & 0.9424 \\
L2R (Dot)-H1 & 0.0105 & 0.9989 \\
L2R (SIPS)-H1 & \textbf{0.1806} & 0.7036 \\
L2R (SIPS)-H16 & 0.1718 & \textbf{0.6701} \\
\bottomrule
\end{tabular}
}
\vspace{-2mm}
\end{wraptable}

To complement PCA visualizations and routing heatmaps, we introduce two quantitative metrics for expert discrimination.
\emph{Mean Routing Margin} measures the average logit gap between the Top-1 expert and the Top-2 expert, while \emph{Low-Margin Rate} measures the proportion of tokens whose routing margin is smaller than $0.2$.
A larger margin indicates clearer separation between the selected expert and its strongest competitor, whereas a lower low-margin rate indicates fewer tokens near the routing decision boundary.
As shown in Table~\ref{tab:expert_discrimination}, X-MoE and L2R (Dot)-H1 exhibit small margins and high low-margin rates, suggesting weak expert separation.
In contrast, L2R with SIPS substantially increases the routing margin and reduces the low-margin rate, providing direct quantitative evidence that SIPS leads to more separated and decisive routing decisions.

\subsection{Routing Robustness under Perturbations}
\label{app:routing_robustness}

\begin{wraptable}{r}{0.48\textwidth}
\vspace{-1.2em}
\centering
\caption{Routing robustness under Gaussian perturbations.}
\label{tab:routing_robustness}
\small
\setlength{\tabcolsep}{5pt}
\renewcommand{\arraystretch}{1.08}
\begin{tabular}{lcc}
\toprule
\textbf{Model} & \textbf{Stability $\uparrow$} & \textbf{Top-$k$ Overlap $\uparrow$} \\
\midrule
Linear & 0.9744 & 0.9679 \\
Linear (SIPS) & \textbf{0.9853} & \textbf{0.9805} \\
X-MoE & 0.9566 & 0.9683 \\
L2R (Dot)-H1 & 0.6734 & 0.8311 \\
L2R (SIPS)-H1 & \textbf{0.9847} & \textbf{0.9915} \\
\bottomrule
\end{tabular}
\begin{minipage}{0.96\linewidth}
\scriptsize
\emph{Note:} Higher values indicate more stable routing decisions.
\end{minipage}
\vspace{-2mm}
\end{wraptable}

To directly validate the stability benefit of SIPS, we apply Gaussian noise
with $\sigma=0.02$ to router inputs and measure two perturbation-based routing
robustness metrics: Stability of the top-1 routed expert and Top-$k$ Overlap (Jaccard similarity of the selected top-8 expert set).

As shown in Table~\ref{tab:routing_robustness}, these results provide direct
empirical support for the stability claim. Comparing Linear to Linear (SIPS)
shows that SIPS improves routing robustness even in the full-rank representation
space, indicating that the scoring function itself contributes to stable expert
selection. More importantly, low-rank routing with unconstrained dot-product
scoring, i.e., L2R (Dot)-H1, becomes highly fragile, with stability dropping to
0.6734 and top-$k$ overlap dropping to 0.8311. In contrast, replacing the
unbounded dot product with SIPS largely restores robustness, achieving 0.9847
stability and 0.9915 top-$k$ overlap for L2R (SIPS)-H1. These results clarify
the component roles: the low-rank projection reshapes the routing geometry,
whereas SIPS makes the resulting low-rank routing stable in practice. We
therefore emphasize bounded-domain control of radial sensitivity, rather than
unconstrained global smoothness.

\section{Additional Geometry of Query-Magnitude Saturation}
\label{sec:appendix_phi_beta}
\begin{figure*}[htbp]
    \centering
    \begin{subfigure}[t]{0.19\linewidth}
        \centering
        \includegraphics[width=\linewidth]{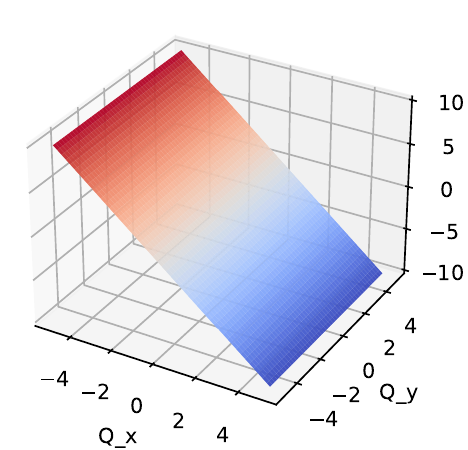}
        \caption{Linear (Dot Product)}
        \label{fig:phi_beta_linear}
    \end{subfigure}
    \hfill
    \begin{subfigure}[t]{0.19\linewidth}
        \centering
        \includegraphics[width=\linewidth]{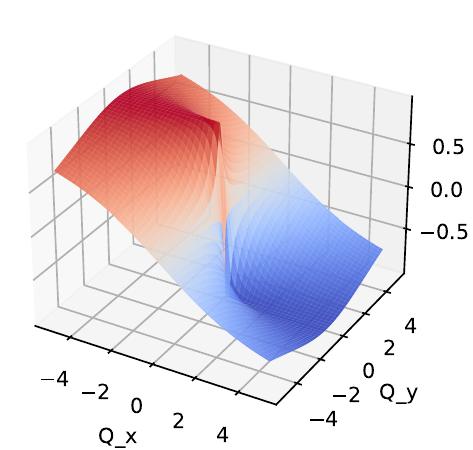}
        \caption{Cosine Scoring}
        \label{fig:phi_beta_cosine}
    \end{subfigure}
    \hfill
    \begin{subfigure}[t]{0.19\linewidth}
        \centering
        \includegraphics[width=\linewidth]{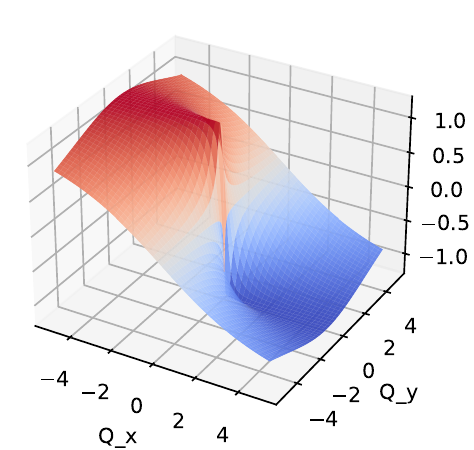}
        \caption{SIPS ($\beta=0.0$)}
        \label{fig:phi_beta_sips_0}
    \end{subfigure}
      \hfill
    \begin{subfigure}[t]{0.19\linewidth}
        \centering
        \includegraphics[width=\linewidth]{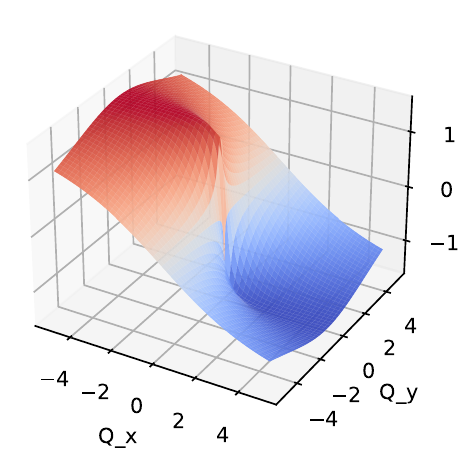}
        \caption{SIPS ($\beta=0.25$)}
        \label{fig:phi_beta_sips_025}
    \end{subfigure}
    \hfill
    \begin{subfigure}[t]{0.19\linewidth}
        \centering
        \includegraphics[width=\linewidth]{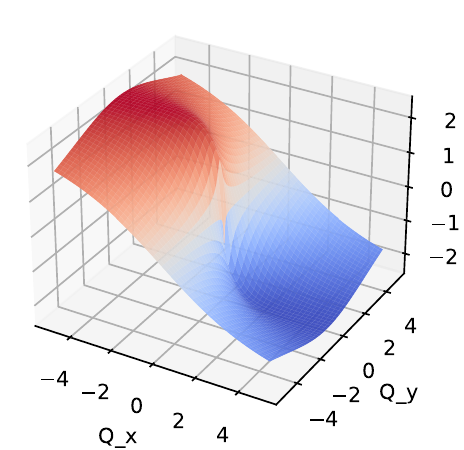}
        \caption{SIPS ($\beta=1.0$)}
        \label{fig:phi_beta_sips_1}
    \end{subfigure}

    \caption{\textbf{Effect of query-magnitude saturation strength ($\beta$) under a fixed anchor.}
    We fix the expert anchor at $\bm{k}=[-2,0]$ and visualize routing logits over query locations $\bm{q}=(Q_x,Q_y)$ in a 2D routing plane (rank $r=2$).
    Dot-product scoring produces an unbounded, nearly half-space-like field dominated by radial growth (a), whereas cosine scoring removes magnitude effects entirely (b).
    SIPS interpolates between these extremes by preserving magnitude as a bounded confidence signal: increasing $\beta$ enlarges the effective dynamic range of the bounded magnitude term while maintaining controlled radial growth (c--e).
    }
    \label{fig:phi_beta_sweep}
\end{figure*}

Figure~\ref{fig:phi_beta_sweep} provides a complementary 3D visualization of routing score landscapes under a fixed anchor, highlighting how different scoring rules respond to variations in query magnitude.
We fix $\bm{k}=[-2,0]$ and sweep $\bm{q}=(Q_x,Q_y)$ over a 2D routing plane, plotting the resulting logit value $z(\bm{q},\bm{k})$.

\paragraph{Dot-Product Scoring.}
As shown in Figure~\ref{fig:phi_beta_linear}, standard dot-product scoring yields an (approximately) half-space-like field whose value grows linearly with $\|\bm{q}\|$ along the radial direction.
This implies that scale variations can dominate the score landscape, creating large regions with weak angular contrast while simultaneously allowing extreme magnitudes to induce overly sharp routing.

\paragraph{Cosine Scoring.}
Cosine scoring (Figure~\ref{fig:phi_beta_cosine}) eliminates magnitude effects by normalizing both $\bm{q}$ and $\bm{k}$, thereby improving global scale stability.
However, the normalization introduces a sharp behavior around $\|\bm{q}\|\approx 0$, where small perturbations of low-norm queries can cause disproportionately large changes in direction (and thus in $\cos\theta$).
Since low-norm queries are also more susceptible to noise, this ``origin sharpness'' can translate into unstable and noisy expert selection.

\paragraph{SIPS Scoring.}
SIPS introduces a bounded query-magnitude transform $\phi(\|\bm{q}\|)=\gamma(1+\beta\tanh(\widehat{\|\bm{q}\|}))$ and thus treats magnitude as a \emph{bounded confidence} factor rather than an unbounded amplifier.
When $\beta=0$ (Figure~\ref{fig:phi_beta_sips_0}), $\phi(\|\bm{q}\|)$ becomes a constant, and the logit reduces to a cosine-like form up to a global rescaling, removing magnitude participation.
As $\beta$ increases (Figure~\ref{fig:phi_beta_sips_025} and~\ref{fig:phi_beta_sips_1}), magnitude information is gradually reintroduced, but its influence remains strictly bounded; importantly, the score landscape becomes noticeably \textbf{smoother around the origin} while preserving \textbf{clear directional selectivity}.
This yields a controlled radial growth together with improved near-origin smoothness, which empirically corresponds to more stable routing dynamics and less noise-sensitive expert selection.

\paragraph{Takeaway.}
Overall, Figure~\ref{fig:phi_beta_sweep} illustrates that SIPS is not merely ``bounded'' in value, but also reshapes the score geometry into a smoother, angle-sensitive landscape.
The saturation strength $\beta$ governs how strongly query magnitude participates in routing: smaller $\beta$ emphasizes angular discrimination with minimal scale effects, while larger $\beta$ leverages magnitude as a bounded confidence signal without recovering the unbounded norm dominance of dot-product scoring.

\section{Gradient Sensitivity and Lipschitz Bounds of SIPS}
\label{sec:appendix_lipschitz}

We analyze the gradient sensitivity of the SIPS logit in Equation~\eqref{eq:sips_factored} and derive Lipschitz bounds
under mild domain assumptions that match practical routing implementations.

\paragraph{Setup.}
For a single expert, define $\rho=\|\bm{q}\|$, $\kappa=\|\bm{k}\|$, $\bm{u}=\bm{q}/\rho$, $\bm{v}=\bm{k}/\kappa$,
and $c=\bm{u}^\top\bm{v}=\cos\theta\in[-1,1]$.
The SIPS logit can be written as:
\begin{equation}
z(\bm{q},\bm{k})=\phi(\rho)\,\psi(\kappa)\,c.
\label{eq:app_sips_core}
\end{equation}
As in practice, we assume numerical stabilization by enforcing:
\begin{equation}
\rho\ge \varepsilon_q,\quad \kappa\ge \varepsilon_k,
\label{eq:app_eps}
\end{equation}
for small $\varepsilon_q,\varepsilon_k>0$.

\paragraph{Cosine gradient bound.}
Using $\bm{u}=\bm{q}/\rho$ and treating $\bm{v}$ as fixed, the cosine alignment satisfies:
\begin{equation}
\nabla_{\bm{q}} c=\frac{1}{\rho}\big(\bm{I}-\bm{u}\bm{u}^\top\big)\bm{v},
\quad
\|\nabla_{\bm{q}} c\|\le \frac{1}{\rho}.
\label{eq:app_grad_c_q}
\end{equation}
Symmetrically:
\begin{equation}
\nabla_{\bm{k}} c=\frac{1}{\kappa}\big(\bm{I}-\bm{v}\bm{v}^\top\big)\bm{u},
\quad
\|\nabla_{\bm{k}} c\|\le \frac{1}{\kappa}.
\label{eq:app_grad_c_k}
\end{equation}

\paragraph{Gradient sensitivity of SIPS.}
Differentiating Equation~\eqref{eq:app_sips_core} yields:
\begin{equation}
\nabla_{\bm{q}} z
=\psi(\kappa)\Big(\phi'(\rho)\,c\,\bm{u}+\phi(\rho)\,\nabla_{\bm{q}} c\Big),
\label{eq:app_grad_z_q}
\end{equation}
\begin{equation}
\nabla_{\bm{k}} z
=\phi(\rho)\Big(\psi'(\kappa)\,c\,\bm{v}+\psi(\kappa)\,\nabla_{\bm{k}} c\Big).
\label{eq:app_grad_z_k}
\end{equation}
Using $|c|\le 1$ and Equations~\eqref{eq:app_grad_c_q}--\eqref{eq:app_grad_c_k}, we obtain:
\begin{equation}
\|\nabla_{\bm{q}} z\|
\le \psi(\kappa)\Big(|\phi'(\rho)|+\frac{\phi(\rho)}{\rho}\Big),
\label{eq:app_grad_bounds1}
\end{equation}
\begin{equation}
\|\nabla_{\bm{k}} z\|
\le \phi(\rho)\Big(|\psi'(\kappa)|+\frac{\psi(\kappa)}{\kappa}\Big).
\label{eq:app_grad_bounds2}
\end{equation}

\paragraph{Lipschitz bounds on a practical domain.}
A global Lipschitz constant requires bounded magnitudes.
We therefore analyze $z$ on the domain:
\begin{equation}
\mathcal{D}=\{\bm{q},\bm{k}:\ \rho\in[\varepsilon_q,\,\rho_{\max}],\ \kappa\in[\varepsilon_k,\,\kappa_{\max}]\},
\label{eq:app_domain}
\end{equation}
where $\rho_{\max}$ is induced by normalization and $\kappa_{\max}$ is induced by weight decay or explicit norm control.
Assume on $\mathcal{D}$ that $\phi(\rho)\le \phi_{\max}$, $|\phi'(\rho)|\le L_\phi$,
and $\psi(\kappa)\le \psi_{\max}$, $|\psi'(\kappa)|\le L_\psi$.
Then by Equations~\eqref{eq:app_grad_bounds1}--\eqref{eq:app_grad_bounds2}, $z$ is Lipschitz in $\bm{q}$ and $\bm{k}$ with:
\begin{equation}
\mathrm{Lip}_{\bm{q}}(z)\le \psi_{\max}\Big(L_\phi+\frac{\phi_{\max}}{\varepsilon_q}\Big),
\end{equation}
\begin{equation}
\mathrm{Lip}_{\bm{k}}(z)\le \phi_{\max}\Big(L_\psi+\frac{\psi_{\max}}{\varepsilon_k}\Big).
\label{eq:app_lipschitz_qk}
\end{equation}
A joint Lipschitz bound with respect to $(\bm{q},\bm{k})$ follows by combining the two gradient bounds.

\paragraph{Instantiation for our transforms.}
For the query transform $\phi(\rho)=\gamma(1+\beta\tanh(\widehat{\rho}))$, we have:
\begin{equation}
\phi_{\max}=\gamma(1+\beta),
\qquad
\Big|\frac{d\phi}{d\widehat{\rho}}\Big|\le \gamma\beta.
\label{eq:app_phi_bounds}
\end{equation}
Thus $L_\phi\le \gamma\beta\,L_{\text{norm}}$ if the scalar normalization map
$\rho\mapsto \widehat{\rho}$ is $L_{\text{norm}}$-Lipschitz on $\mathcal{D}$.
For the anchor transform $\psi(\kappa)=1+(\kappa-1)/p$, we have:
\begin{equation}
L_\psi=\frac{1}{p},
\qquad
\psi_{\max}=1+\frac{\kappa_{\max}-1}{p}\ \ \text{on}\ \mathcal{D}.
\label{eq:app_psi_bounds}
\end{equation}

\paragraph{Contrast to dot-product scoring.}
For $z_{\text{dot}}(\bm{q},\bm{k})=\bm{q}^\top\bm{k}$, the gradients are
$\nabla_{\bm{q}} z_{\text{dot}}=\bm{k}$ and $\nabla_{\bm{k}} z_{\text{dot}}=\bm{q}$,
so sensitivity grows linearly with $\|\bm{k}\|$ and $\|\bm{q}\|$.
In contrast, SIPS bounds the radial sensitivity via bounded/contractive magnitude transforms,
and therefore yields an explicit mechanism to control the Lipschitz behavior of the scoring map.

\section{Complexity and Efficiency}
\label{sec:efficiency}

We analyze the routing overhead of L2R and compare it to the standard linear router.
Let $d$ be the backbone hidden dimension, $N$ the number of experts, $r$ the routing rank ($r\ll d$), $k$ the top-$k$ sparsity, and $H$ the number of anchors per expert in multi-anchor heads.

\paragraph{Baseline: Linear Routing.}
A standard linear router computes logits by
$\bm{z}=\bm{x}\bm{W}_g$ with $\bm{W}_g\in\mathbb{R}^{d\times N}$ as Equation~\eqref{eql:linear}.
The routing parameter count is:
\begin{equation}
    P_{\text{lin}} = dN,
\end{equation}
and the per-token routing cost is dominated by the matrix-vector product:
\begin{equation}
    C_{\text{lin}} \approx O(dN).
\end{equation}

\paragraph{L2R Routing.}
L2R replaces the high-dimensional projection by (i) a shared low-rank projection $\bm{W}_q\in\mathbb{R}^{d\times r}$ and
(ii) expert anchors $\{\bm{k}_{i,h}\}$ in $\mathbb{R}^{r}$.
With multi-anchor heads, L2R maintains $NH$ anchors.

\textbf{Parameters.}
The routing parameters consist of $\bm{W}_q$ and anchors:
\begin{equation}
    P_{\text{L2R}} = dr + NHr.
\end{equation}
Thus, the parameter reduction over linear routing is:
\begin{equation}
    \frac{P_{\text{L2R}}}{P_{\text{lin}}} 
    = \frac{dr+NHr}{dN}
    = \frac{r}{N} + \frac{Hr}{d}.
\end{equation}
In typical MoE settings where $d\gg r$ and $N$ is moderate-to-large, the router overhead is substantially reduced, and the multi-anchor term remains small because it scales only with $r$.

\textbf{Computation.}
L2R first computes $\bm{q}=\bm{x}\bm{W}_q$ with cost $O(dr)$, then evaluates anchor logits by low-rank inner products with cost $O(NHr)$:
\begin{equation}
    C_{\text{L2R}} \approx O(dr + NHr).
\end{equation}
SIPS adds only lightweight scalar transforms on norms and does not change the asymptotic complexity.
Multi-anchor pooling aggregates $H$ logits per expert as Equation~\eqref{eq:multi_anchor_pool}, which is $O(NH)$ and negligible compared to $O(NHr)$ when $r\ge 2$.

\paragraph{Measured Router Parameters on OLMoE.}
Table~\ref{tab:router_params_grid} reports router parameter counts under different rank--head combinations on OLMoE (16 MoE layers, $d{=}2048$, $N{=}64$).
Each entry shows the total router parameters aggregated over all MoE layers, followed by the percentage relative to the Linear router (2.097M parameters).
Across all tested configurations, router parameters remain negligible compared to the full model size ($\sim$6.92B), and multi-anchor heads incur only a minor additional cost.

\begin{table}[htbp]
\centering
\caption{\textbf{Router parameters under different rank $r$ and head $H$ settings on OLMoE.}
Each cell reports total router params over all MoE layers, with the percentage relative to the Linear router (2.097M) in parentheses.}
\label{tab:router_params_grid}
\scriptsize
\setlength{\tabcolsep}{4pt}
\begin{tabular}{c|ccccc}
\toprule
Rank $r$ $\backslash$ Heads $H$ & 1 & 2 & 4 & 8 & 16 \\
\midrule
2  & 100.352K (4.79\%) & 102.400K (4.88\%) & 106.496K (5.08\%) & 114.688K (5.47\%) & 131.072K (6.25\%) \\
4  & 167.936K (8.01\%) & 172.032K (8.20\%) & 180.224K (8.59\%) & 196.608K (9.38\%) & 229.376K (10.94\%) \\
8  & 303.104K (14.46\%) & 311.296K (14.84\%) & 327.680K (15.62\%) & 360.448K (17.19\%) & 425.984K (20.32\%) \\
16 & 573.440K (27.34\%) & 589.824K (28.12\%) & 622.592K (29.69\%) & 688.128K (32.81\%) & 819.200K (39.07\%) \\
32 & 1.114M (53.11\%) & 1.147M (54.68\%) & 1.212M (57.81\%) & 1.343M (64.06\%) & 1.606M (76.59\%) \\
\bottomrule
\end{tabular}
\end{table}

\paragraph{Discussion: FLOPs vs.\ Wall-Clock Time.}
Although $O(dr+NHr)$ can be significantly smaller than $O(dN)$, practical end-to-end speedups are not guaranteed.
Routing is often not the dominant cost of MoE blocks, and the L2R router introduces additional element-wise operations 
(normalization, $\tanh$, and pooling) that may be less hardware-efficient than a single dense GEMM.

To quantify practical inference cost, we further measure wall-clock latency and throughput on 8$\times$B200 GPUs with batch size 1, averaged over 5 samples.

\begin{table}[htbp]
\centering
\caption{Inference overhead on OLMoE.}
\label{tab:inference_overhead}
\small
\setlength{\tabcolsep}{5pt}
\renewcommand{\arraystretch}{1.05}
\begin{tabular}{lcccc}
\toprule
Model & Prefill Latency (ms) $\downarrow$ & Prefill Tok/s $\uparrow$ & Decode Latency (ms/step) $\downarrow$ & Decode Tok/s $\uparrow$ \\
\midrule
Linear & 180.78 & 22657.39 & 43.79 & 22.84 \\
L2R (Dot)-H1 & 181.94 & 22512.97 & 46.29 & 21.60 \\
L2R (SIPS)-H1 & 187.08 & 21894.31 & 47.06 & 21.25 \\
L2R (SIPS)-H16 & 182.36 & 22461.40 & 47.12 & 21.22 \\
\bottomrule
\end{tabular}
\vspace{-2mm}
\end{table}

As shown in Table~\ref{tab:inference_overhead}, L2R introduces only a small prefill overhead relative to Linear.
For example, L2R (SIPS)-H16 is close to Linear in prefill latency 
(182.36 ms vs. 180.78 ms), while decode latency increases by about 3.3 ms/step.
This is consistent with our complexity analysis: L2R is parameter-efficient, but theoretical savings do not automatically translate into end-to-end wall-clock gains without further kernel- and system-level optimization.
\section{Optimization Objective Details}
\label{sec:objective_details}

We train L2R under the standard optimization objective adopted by prior sparse MoE models, without introducing additional loss terms.
Specifically, the objective consists of the task loss, a load-balancing loss to encourage uniform expert utilization, and the router z-loss~\citep{zoph2022stmoe} used in OLMoE~\citep{muennighoff2025olmoe} to stabilize gating logits in large-scale language model training.

\paragraph{Load-Balancing Loss.}
Following prior MoE work, we employ a load-balancing loss to prevent expert under-utilization and collapse.
Let $s_{t,i}$ denote the routing probability assigned to expert $i$ for token $t$, and let $\mathbb{I}[i \in \mathcal{T}_t]$ indicate whether expert $i$ is selected in the top-$k$ routing set for token $t$.
Over a batch of $T$ tokens, we define the average routing probability and empirical expert usage as:
\begin{equation}
    \bar{s}_i = \frac{1}{T}\sum_{t=1}^{T} s_{t,i},
    \qquad
    f_i = \frac{1}{T}\sum_{t=1}^{T} \mathbb{I}[i \in \mathcal{T}_t].
\end{equation}
The load-balancing loss is then given by:
\begin{equation}
    \mathcal{L}_{\text{bal}} = N \sum_{i=1}^{N} \bar{s}_i\, f_i,
\end{equation}
which encourages alignment between routing probabilities and actual expert assignments.

\paragraph{Z-Loss.}
Following OLMoE~\citep{muennighoff2025olmoe}, we additionally employ the router z-loss for large-scale language model training to regularize the magnitude of pre-softmax routing logits.
Let $\bm{z}_t \in \mathbb{R}^{N}$ denote the routing logits for token $t$.
The z-loss is defined as:
\begin{equation}
    \mathcal{L}_{z}
    = \frac{1}{T}\sum_{t=1}^{T}
    \Big(\log \sum_{i=1}^{N} \exp(z_{t,i})\Big)^2,
\end{equation}
which discourages excessively large logits and improves training stability.
We apply this term only in LLM training, and omit it for vision experiments.

\paragraph{Overall Objective.}
The final optimization objective is:
\begin{equation}
    \mathcal{L}
    = \mathcal{L}_{\text{task}}
    + \lambda_{\text{bal}}\,\mathcal{L}_{\text{bal}}
    + \lambda_{z}\,\mathcal{L}_{z},
\end{equation}
where $\lambda_{\text{bal}}$ and $\lambda_{z}$ control the relative weights of the auxiliary losses.

\section{Experiments with OLMoE: Data Preparation}
\label{app:olmoe_subset}
\begin{table}[htbp]
\centering
\caption{\textbf{Distribution-matched 200B subset of \texttt{allenai/OLMoE-mix-0924}.}
We randomly sample complete shards per domain with a fixed seed (2025) to obtain a \(\approx\)200B-token subset while preserving the domain mixture ratios of the full dataset.}
\label{tab:olmoe_subset_dist}
\begin{small}
\begin{tabular}{lrrrr}
\toprule
\textbf{Domain} & \textbf{Total shards} & \textbf{Selected shards} & \textbf{Approx.\ tokens (B)} & \textbf{Token share (\%)} \\
\midrule
dclm            & 1970 & 97 & 190.06 & 94.01 \\
starcoder       &  863 & 43 &   5.03 &  2.49 \\
pes2o           &   26 &  1 &   2.20 &  1.09 \\
open-web-math   &   26 &  1 &   0.49 &  0.24 \\
algebraic-stack &   16 &  1 &   0.79 &  0.39 \\
wiki            &    2 &  1 &   1.84 &  0.91 \\
\midrule
\textbf{Total}  & 2903 & 144 & 200.41 & 100.00 \\
\bottomrule
\end{tabular}
\end{small}
\end{table}

\paragraph{Source dataset.}
We use the official OLMoE pretraining mixture \texttt{allenai/OLMoE-mix-0924}, which contains approximately 4.07T tokens spanning six domains.
To accelerate experimentation while preserving the training distribution, we construct a smaller, distribution-matched subset by sampling complete shards from each domain in proportion to its scale in the full mix.

\paragraph{Distribution-matched shard sampling.}
Let $\mathcal{D}=\{d\}$ denote the set of domains and let $S_d$ be the number of shards in domain $d$.
Given a target budget of $T$ tokens (e.g., $T\!=\!200$B), we randomly sample a fixed number of shards from each domain using a fixed seed (seed=2025), such that the resulting token mass closely matches the original mixture proportions.
Concretely, we sample shards without replacement within each domain, and keep the domain-wise sampling ratios aligned with the full dataset scale.
Table~\ref{tab:olmoe_subset_dist} reports the resulting domain composition for the 200B subset used in our main OLMoE experiments.

\paragraph{Multiple subset sizes.}
Using the same procedure, we also generate additional subsets at smaller scales (10B/50B/100B tokens) for fast ablations and debugging.
All subsets are constructed by the same distribution-matched shard sampling strategy to ensure comparability across scales.

\section{Training Configuration Details}\label{sec:appendix_config}
\subsection{Training Configuration for OLMoE Experiments}
\label{sec:appendix_olmoe_config}
This appendix reports the core training hyperparameters for OLMoE experiments in Table~\ref{tab:olmoe_hparams}.
To isolate routing effects, we follow the original OLMoE~\citep{muennighoff2025olmoe} settings and keep the backbone architecture, optimization recipe, and MoE auxiliary losses fixed across all router variants, modifying only router-specific components (e.g., projection rank, scoring, or head design) when applicable.

\begin{table}[htbp]
\centering
\small
\setlength{\tabcolsep}{6pt}
\renewcommand{\arraystretch}{1.05}
\caption{\textbf{Key hyperparameters for OLMoE training.}}
\label{tab:olmoe_hparams}
\resizebox{0.9\linewidth}{!}{
\begin{tabular}{p{0.19\linewidth}p{0.1\linewidth}
                p{0.21\linewidth}p{0.1\linewidth}
                p{0.18\linewidth}p{0.1\linewidth}}
\toprule
\textbf{Item}  & \textbf{Value} &
\textbf{Item}  & \textbf{Value} &
\textbf{Item}  & \textbf{Value} \\
\midrule
$d_{\text{model}}$ & 2,048 &
 Max sequence length & 4,096 &
Experts $N$ & 64 \\
 Transformer layers $L$ & 16 &
Vocabulary size & 50,280 &
Top-$k$ & 8 \\
 Attention heads $n_{\text{heads}}$ & 16 &
 Load-balance weight $\lambda_{\text{bal}}$ & 0.01 &
 Dropless routing & on \\
MLP ratio & 1 &
Z-loss weight $\lambda_{z}$ & 0.001 &
Type & AdamW \\
 Activation & SwiGLU &
 Learning rate & $4\times10^{-4}$ &
 $(\beta_1,\beta_2)$ & $(0.9,0.95)$ \\
 Norm & RMSNorm &
 $\epsilon$ & $10^{-8}$ &
Weight decay & 0.1 \\
 Warmup tokens $t_{\text{warmup}}$ & $3\times10^{8}$ &
 Total tokens $t_{\max}$ & $10^{10}$ &
Decay & cosine \\
Global batch size & 1024 &
Microbatch size & 1 &
 Distributed training & DDP \\
\bottomrule
\end{tabular}}
\vspace{-2mm}
\end{table}

\paragraph{Implementation Details.}
All models are trained from scratch on the 10B-token in-distribution dataset described in Appendix~\ref{app:olmoe_subset}.
We follow the standard OLMoE recipe with a Transformer backbone of $d_{\text{model}}=2048$, $N=64$ experts per MoE block, and top-$k=8$ sparse activation.
Optimization is performed using AdamW with cosine learning-rate decay, and all experiments are trained with Distributed Data Parallel (DDP) for scalability and stability.
All runs are conducted on an $8\times$ NVIDIA B200 (Blackwell) setup.

The original OLMoE implementation relies on \texttt{Megablocks}~\citep{gale2022megablocks} for MoE kernel acceleration.
However, \texttt{Megablocks} currently does not support Blackwell \texttt{sm\_100} image configuration, which limits its compatibility on B200.
To ensure correct execution and efficient expert dispatch, we replace the \texttt{Megablocks} MoE module with \texttt{Tutel}~\citep{hwang2023tutel}, which offers architecture-agnostic kernels and stable performance under modern GPU backends.
This change affects only the MoE dispatch backend and does not alter the routing logic or model formulation, ensuring fair and controlled comparisons across routing variants.
For full reproducibility, we will release the complete configuration files and training scripts with the codebase.

\subsection{Training Configuration for ViT Experiments}
\label{sec:appendix_vit_config}
This appendix reports the core training hyperparameters for ViT ImageNet experiments in Table~\ref{tab:vit_hparams}.
To isolate routing effects, we keep the backbone architecture, augmentation recipe, and optimization settings fixed across router variants, and only modify router-specific components (e.g., router type, rank, scoring mode, or head design) when applicable.

\begin{table}[t]
\centering
\small
\setlength{\tabcolsep}{6pt}
\renewcommand{\arraystretch}{1.05}
\caption{\textbf{Key hyperparameters for ViT-S ImageNet-1K training.}}
\label{tab:vit_hparams}
\resizebox{0.95\linewidth}{!}{
\begin{tabular}{p{0.19\linewidth}p{0.1\linewidth}
                p{0.21\linewidth}p{0.1\linewidth}
                p{0.18\linewidth}p{0.1\linewidth}}
\toprule
\textbf{Item}  & \textbf{Value} &
\textbf{Item}  & \textbf{Value} &
\textbf{Item}  & \textbf{Value} \\
\midrule
Model & ViT-S/16 &
Image size & 224 &
\#Classes & 1000 \\
ViT depth & 12 &
Epochs & 300 &
Batch size (per GPU) & 128 \\
Optimizer & AdamW &
Learning rate & $5\times10^{-4}$ &
Min LR & $10^{-6}$ \\
Weight decay & 0.05 &
Warmup epochs & 5 &
Grad clip & 1.0 \\
RandAugment & on &
Random erasing prob. & 0.25 &
Label smoothing & 0.1 \\
Mixup & 0.8 &
CutMix & 1.0 &
\hphantom{a} & \hphantom{a} \\
\midrule
Experts $N$ & 16 &
Top-$k$ & 2 &
MoE last-$n$ blocks & 4 \\
Aux. weight $\lambda_{\text{bal}}$ & 0.01 &
Learnable $\tau$ & false&
\hphantom{a} & \hphantom{a} \\
\bottomrule
\end{tabular}}
\vspace{-2mm}
\end{table}


\section{Full Experimental Results for OLMoE Experiments}
\label{app:full_result_olmoe}

\begin{table}[htbp]
\centering
\caption{Downstream performance at the checkpoint closest to 10B total tokens (\%). We report all four MMLU groups and their mean (MMLU), four additional tasks, and the overall mean (Overall) across eight tasks. Best results are in bold, and second-best results are underlined.}
\label{tab:downstream_10b_mmlu_breakdown}
\resizebox{\linewidth}{!}{%
\begin{tabular}{l l c|cccc|c|cccc|c}
\toprule
Method & H & Rank & Hum & Other & SocSci & STEM & MMLU & BoolQ & HSwag & PIQA & SciQ & Overall \\
\midrule
X-MoE & - & 32 & 23.4 & \textbf{30.1} & 25.8 & 19.1 & 24.6 & 60.0 & \underline{40.6} & \textbf{64.5} & 68.8 & 41.5 \\
Linear & - & - & 22.3 & 21.5 & 27.0 & 19.9 & 22.7 & 60.0 & 38.7 & 61.1 & 66.8 & 39.6 \\
L2R (Cosine) & - & 2 & 22.3 & 27.3 & 30.1 & 19.9 & 24.9 & 61.1 & \textbf{42.2} & 60.0 & \textbf{72.7} & \underline{42.0} \\
L2R (Dot) & - & 2 & \underline{25.4} & \underline{30.9} & 21.1 & \underline{20.3} & 24.4 & 59.2 & 39.5 & 60.5 & 64.8 & 40.3 \\
\midrule
L2R (SIPS) & 16 & 2 & \textbf{28.9} & 25.8 & \textbf{34.8} & \textbf{21.1} & \textbf{27.6} & 61.7 & \underline{40.6} & 62.3 & 71.5 & \textbf{43.3} \\
L2R (SIPS) & 8 & 2 & 22.7 & 26.2 & \underline{32.0} & 19.9 & 25.2 & 60.2 & 39.8 & \underline{64.3} & 68.4 & 41.7 \\
L2R (SIPS) & 4 & 2 & 24.2 & 28.1 & 29.3 & 19.1 & 25.2 & \underline{62.1} & 39.5 & 61.1 & 69.1 & 41.6 \\
L2R (SIPS) & 4 & 32 & 23.8 & 27.7 & 31.2 & 19.5 & \underline{25.6} & 59.4 & 39.1 & 61.1 & \underline{71.9} & 41.7 \\
L2R (SIPS) & 1 & 2 & 22.7 & 27.0 & 31.2 & \textbf{21.1} & 25.5 & 60.0 & 37.5 & 60.5 & 69.9 & 41.2 \\
L2R (SIPS) & 4 & 8 & 25.0 & 23.8 & 24.6 & 19.5 & 23.2 & \textbf{62.3} & 38.3 & 63.1 & 68.8 & 40.7 \\
\bottomrule
\end{tabular}}
\end{table}

\section{Downstream Evaluation Protocol}
\label{app:eval_protocol}

We follow the downstream evaluation protocol used by {OLMoE}~\citep{muennighoff2025olmoe} and {OLMES}~\citep{gu2025olmes} for reporting model quality.
Specifically, we evaluate a fixed set of commonsense and knowledge benchmarks at checkpoints along training and report the average score (\textbf{Overall}) across all included tasks.
All evaluations are performed using the same prompt format and normalization conventions as in the OLMoE/{OLMES} setup.

\subsection{Tasks and Metrics}
\label{app:eval_tasks_metrics}

Table~\ref{tab:eval_setup} summarizes the tasks and evaluation settings used in this work.
We report task accuracy for all benchmarks.
For MMLU~\citep{hendrycks2021measuring}, we follow the standard multiple-choice formulation and report both the mean MMLU score and the four group scores (Humanities, Social Sciences, STEM, Other) in Appendix~\ref{app:full_result_olmoe}.

\begin{table}[htbp]
\centering
\caption{\textbf{Summary of downstream evaluation.} We adopt the OLMoE/{OLMES} evaluation conventions.
CF = completion/cloze (generative) formulation; MCF = multiple-choice formulation;
Norm indicates the answer normalization used by the evaluator.}
\label{tab:eval_setup}
\small
\setlength{\tabcolsep}{4pt}
\begin{tabular}{lccc}
\toprule
\textbf{Dataset} & \textbf{Format} & \textbf{Norm} & \textbf{Metric} \\
\midrule
MMLU~\citep{hendrycks2021measuring} & MCF (5-shot) & char & Accuracy \\
BoolQ~\citep{clark2019boolq} & CF (0-shot) & none & Accuracy \\
HellaSwag~\citep{zellers2019hellaswag} & CF (0-shot) & char & Accuracy \\
PIQA~\citep{Bisk2019PIQARA} & CF (0-shot) & char & Accuracy \\
SciQ~\citep{welbl2017crowdsourcing} & CF (0-shot) & none & Accuracy \\
\bottomrule
\end{tabular}
\end{table}

\subsection{Early-Stage Evaluation}
\label{app:eval_excluded}

Our language MoE models are trained from scratch up to \textbf{10B tokens}.
At this early pretraining stage, some downstream benchmarks commonly reported in OLMoE/{OLMES} (e.g., {WinoGrande}~\citep{sakaguchi2019winogrande} and ARC-Challenge~\citep{clark2018think}) tend to exhibit high variance and unstable ranking across checkpoints, because the model has not yet formed sufficiently robust linguistic and commonsense representations.
Including such tasks at 10B tokens can introduce noise that obscures routing-induced differences.
Therefore, we focus on a compact set of tasks (Table~\ref{tab:eval_setup}) that are more stable and informative in the 10B-token regime, and we report an overall mean across these tasks for comparability.

\subsection{Benchmark Descriptions}
\label{app:benchmark_descriptions}

We evaluate downstream quality using a compact suite of widely-used benchmarks that probe complementary capabilities (knowledge, reading comprehension, commonsense reasoning, and scientific QA).
Below, we briefly describe each benchmark and the associated evaluation interface used in our pipeline.

\paragraph{MMLU.}
Massive Multitask Language Understanding (MMLU) is a broad knowledge benchmark covering dozens of subjects spanning the humanities, social sciences, STEM, and other domains.
Each example is a multiple-choice question with four candidate options.
We evaluate MMLU in the standard multiple-choice formulation (MCF) and report accuracy.
In addition to the overall MMLU mean, we optionally report the four group means (Hum, SocSci, STEM, Other) to diagnose capability skew across domains.

\paragraph{BoolQ.}
BoolQ is a yes/no question answering benchmark grounded in short passages.
Given a passage and a question, the model predicts a binary label (\texttt{yes} or \texttt{no}).
We evaluate with accuracy.
BoolQ emphasizes reading comprehension and entailment-style reasoning, and it is typically less sensitive to prompt formatting than many open-ended generation tasks.

\paragraph{HellaSwag.}
HellaSwag is a commonsense inference benchmark framed as sentence completion.
Given a context, the model must choose the most plausible continuation from multiple candidates.
We compute a score for each candidate continuation via conditional likelihood and report accuracy.
HellaSwag probes grounded commonsense and causal/temporal plausibility, and is commonly used to reflect progress in event-level reasoning.

\paragraph{PIQA.}
PIQA tests physical commonsense reasoning for everyday situations.
Each example describes a goal (e.g., how to accomplish a household task) and provides two candidate solutions; the model must select the more plausible one.
We evaluate via candidate scoring and report accuracy.
PIQA complements HellaSwag by focusing specifically on physical affordances and action feasibility.

\paragraph{SciQ.}
SciQ is a science QA dataset consisting of multiple-choice questions derived from science exams and accompanying explanations.
We evaluate in the standard candidate-scoring setup and report accuracy.
SciQ provides an additional view of structured scientific knowledge.

\clearpage
\section{Visualizations for OLMoE}
\subsection{Visualizations for OLMoE Training Results}
\begin{figure*}[htbp]
    \centering

    \begin{subfigure}[t]{0.19\textwidth}
        \centering
        \includegraphics[width=\linewidth]{figures/plot_mmlu_mean.pdf}

    \end{subfigure}
    \hfill
    \begin{subfigure}[t]{0.19\textwidth}
        \centering
        \includegraphics[width=\linewidth]{figures/plot_hellaswag.pdf}
    \end{subfigure}
    \hfill
    \begin{subfigure}[t]{0.19\textwidth}
        \centering
        \includegraphics[width=\linewidth]{figures/plot_c4_ce.pdf}

    \end{subfigure}
    \hfill
    \begin{subfigure}[t]{0.19\textwidth}
        \centering
        \includegraphics[width=\linewidth]{figures/plot_train_ce.pdf}
    \end{subfigure}
    \hfill
    \begin{subfigure}[t]{0.19\textwidth}
        \centering
        \includegraphics[width=\linewidth]{figures/plot_load_balance.pdf}
    \end{subfigure}

    \begin{subfigure}[t]{0.19\textwidth}
        \centering
        \includegraphics[width=\linewidth]{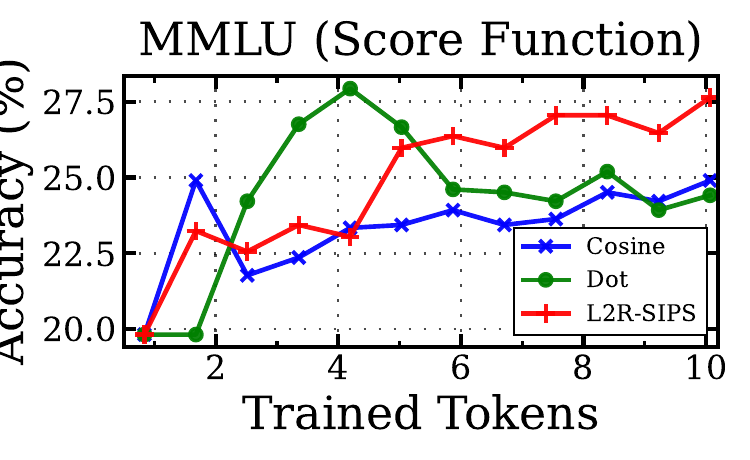}
    \end{subfigure}
    \hfill
    \begin{subfigure}[t]{0.19\textwidth}
        \centering
        \includegraphics[width=\linewidth]{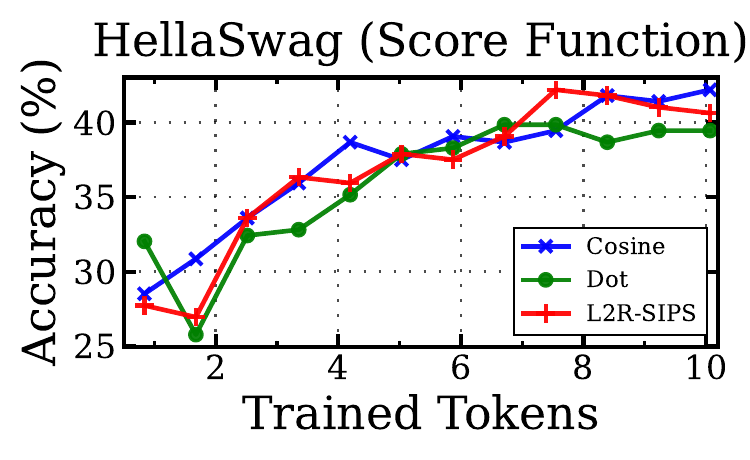}
    \end{subfigure}
    \hfill
    \begin{subfigure}[t]{0.19\textwidth}
        \centering
        \includegraphics[width=\linewidth]{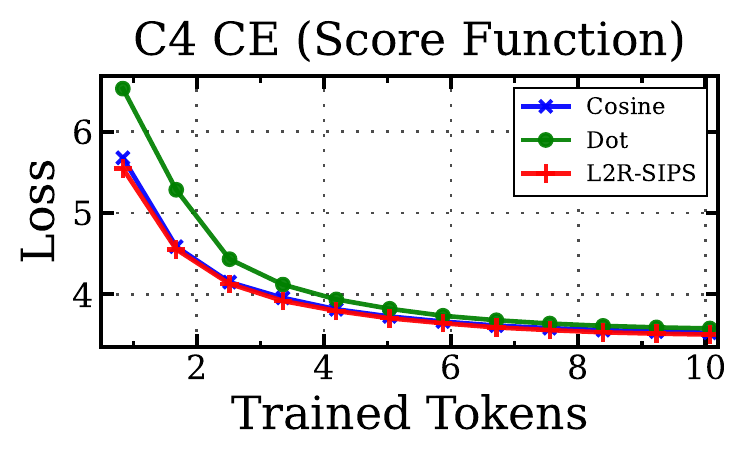}
    \end{subfigure}
    \hfill
    \begin{subfigure}[t]{0.19\textwidth}
        \centering
        \includegraphics[width=\linewidth]{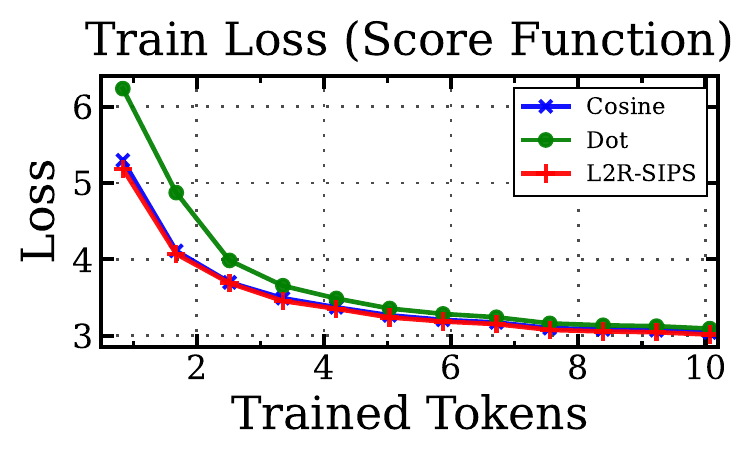}
    \end{subfigure}
    \hfill
    \begin{subfigure}[t]{0.19\textwidth}
        \centering
        \includegraphics[width=\linewidth]{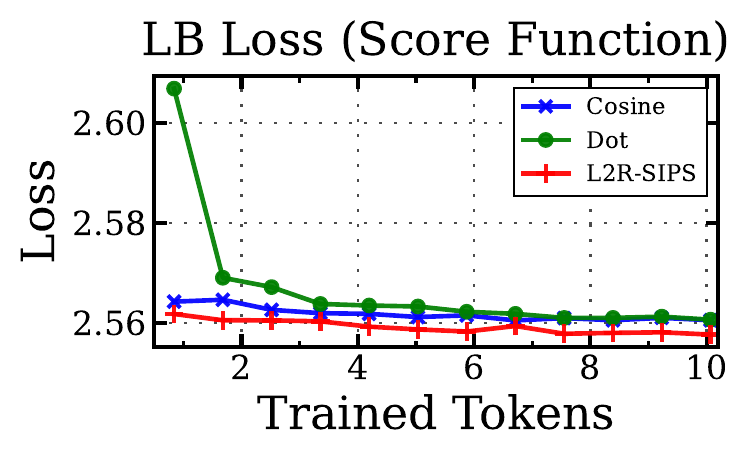}
    \end{subfigure}

        \begin{subfigure}[t]{0.19\textwidth}
        \centering
        \includegraphics[width=\linewidth]{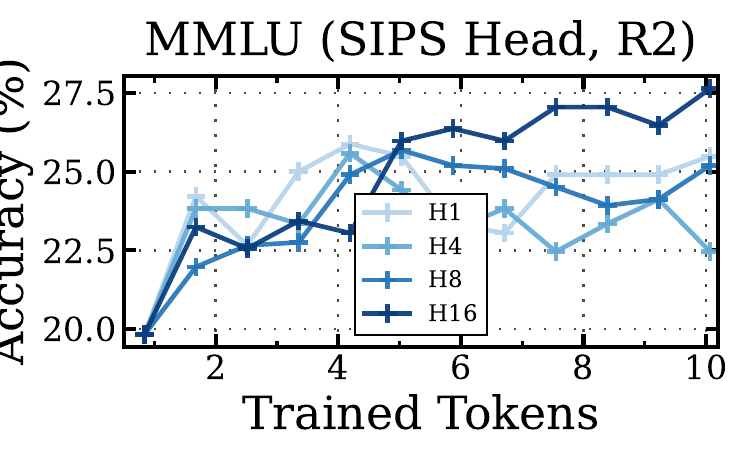}
    \end{subfigure}
    \hfill
    \begin{subfigure}[t]{0.19\textwidth}
        \centering
        \includegraphics[width=\linewidth]{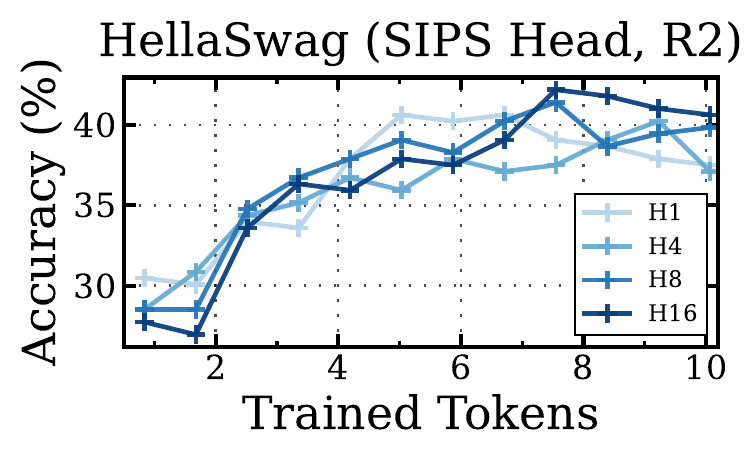}
    \end{subfigure}
    \hfill
    \begin{subfigure}[t]{0.19\textwidth}
        \centering
        \includegraphics[width=\linewidth]{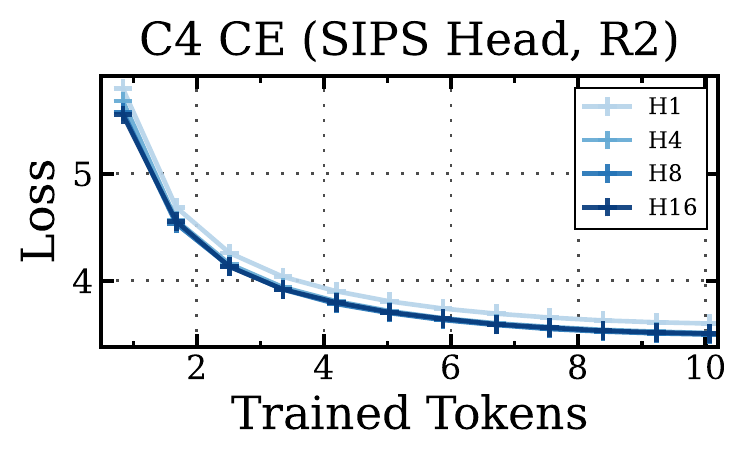}
    \end{subfigure}
    \hfill
    \begin{subfigure}[t]{0.19\textwidth}
        \centering
        \includegraphics[width=\linewidth]{figures/ablation_head_train_loss.pdf}
    \end{subfigure}
    \hfill
    \begin{subfigure}[t]{0.19\textwidth}
        \centering
        \includegraphics[width=\linewidth]{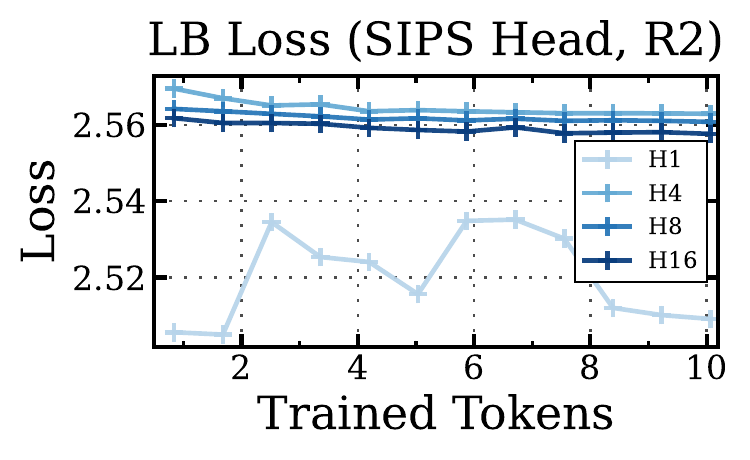}
    \end{subfigure}

    \begin{subfigure}[t]{0.19\textwidth}
        \centering
        \includegraphics[width=\linewidth]{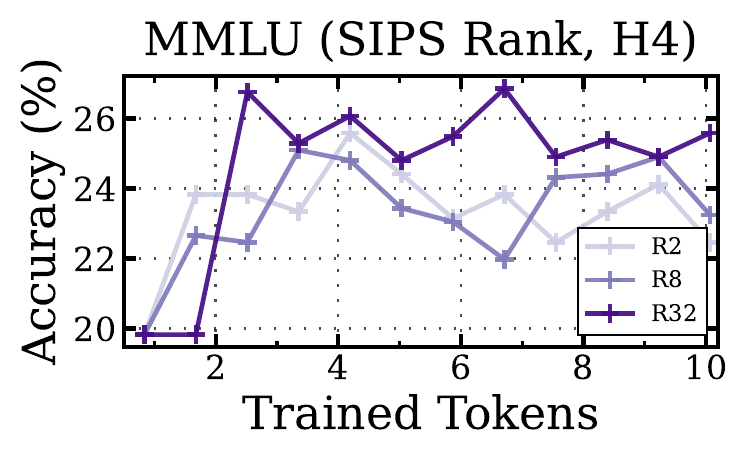}
    \end{subfigure}
    \hfill
    \begin{subfigure}[t]{0.19\textwidth}
        \centering
        \includegraphics[width=\linewidth]{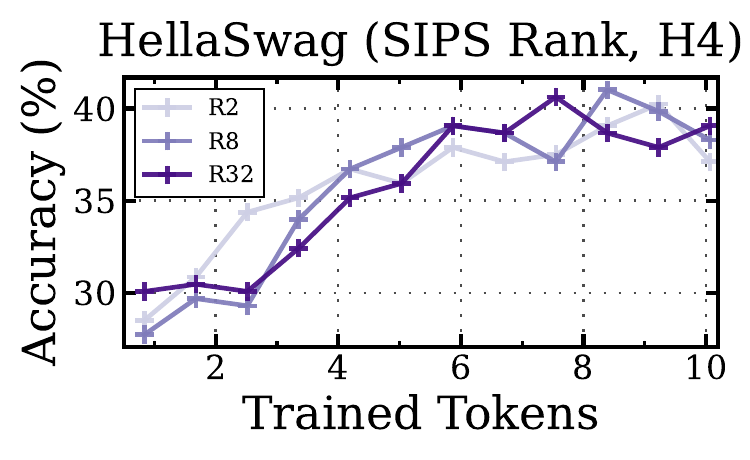}
    \end{subfigure}
    \hfill
    \begin{subfigure}[t]{0.19\textwidth}
        \centering
        \includegraphics[width=\linewidth]{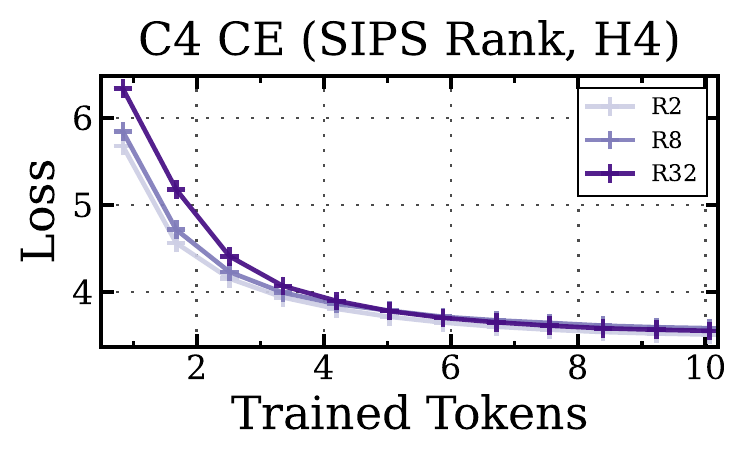}
    \end{subfigure}
    \hfill
    \begin{subfigure}[t]{0.19\textwidth}
        \centering
        \includegraphics[width=\linewidth]{figures/ablation_rank_train_loss.pdf}
    \end{subfigure}
    \hfill
    \begin{subfigure}[t]{0.19\textwidth}
        \centering
        \includegraphics[width=\linewidth]{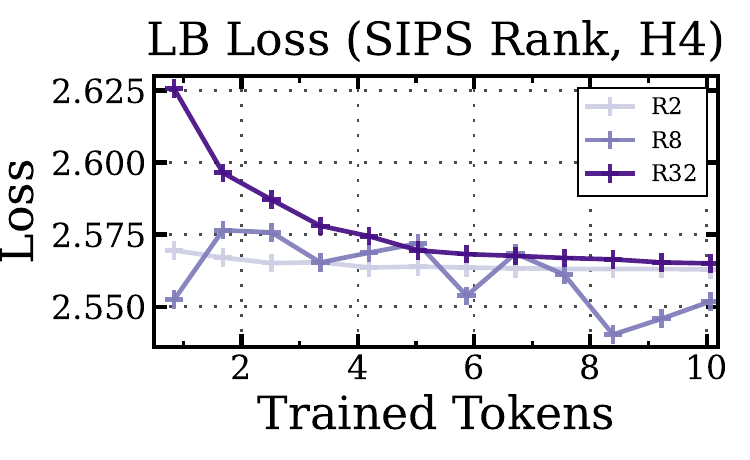}
    \end{subfigure}
    
\vspace{-3mm}
\caption{\textbf{Visualizations of OLMoE training dynamics.}
The panels report MMLU accuracy, HellaSwag accuracy, C4 validation cross-entropy (CE), training CE, and load-balance loss across router variants. From top to bottom, we compare alternative methods, scoring modes, and ablations over head and rank settings. The x-axis denotes the number of trained tokens.}
\label{fig:vvv}
\vspace{3mm}
\end{figure*}

\subsection{Routing Geometry Visualizations in OLMoE}
\label{sec:app_routing_geometry}

\begin{figure*}[htbp]
\vspace{4mm}
    \centering
    \begin{subfigure}[t]{0.3\textwidth}
        \centering
        \includegraphics[width=\linewidth]{\detokenize{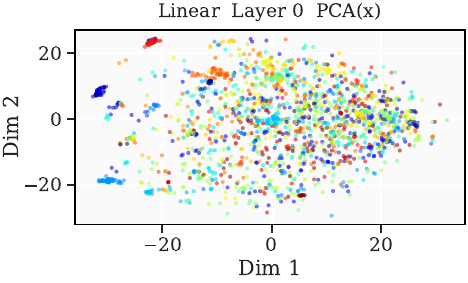}}
    \end{subfigure}
\hfill
    \begin{subfigure}[t]{0.3\textwidth}
        \centering
        \includegraphics[width=\linewidth]{\detokenize{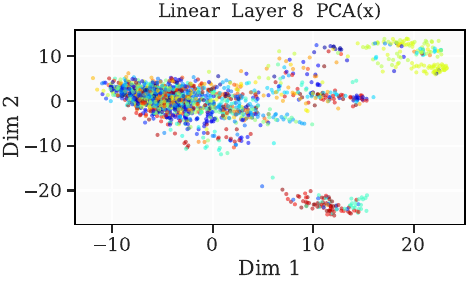}}
    \end{subfigure}
\hfill
    \begin{subfigure}[t]{0.3\textwidth}
        \centering
        \includegraphics[width=\linewidth]{\detokenize{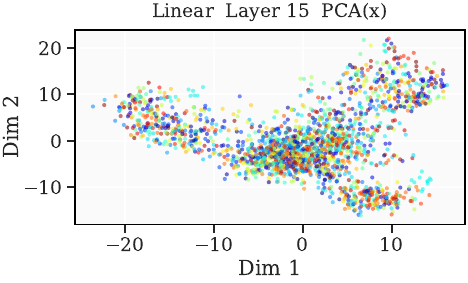}}
    \end{subfigure}

    \caption{\textbf{Linear-router baselines.} PCA scatter plots in backbone representation space
    ($x$-space) for the same layers as Figure~\ref{fig:app_l2r_geometry}.
    Tokens are colored by the top-1 expert assignment under a linear router.
    These baselines help contrast expert separation in raw representations against our L2R latent
    routing space.}
    \label{fig:app_linear_geometry}
\end{figure*}

This section provides qualitative visualizations of routing geometry in the OLMoE setting.
For selected MoE FFN layers (early-0; middle-8; late-15), we project token representations to two
dimensions using PCA\@.
Each point corresponds to a token representation and is colored by its top-1 routed expert.
We visualize (i) the {backbone representation space} ($x$-space) and (ii) the {routing latent space}
($q$-space) produced by our low-rank projection.
These plots support two observations: (1) routing in raw high-dimensional representations can
lead to ambiguous expert separation, while (2) our routing latent space yields clearer cluster structure aligned with expert assignment, making the routing geometry more interpretable.
We additionally report linear-router and X-MoE baselines for direct comparison.

\begin{figure}[p]
\centering
\begin{subfigure}[t]{0.3\textwidth}
\centering
\includegraphics[width=\linewidth]{\detokenize{figures/pca\_ours\_transformer\_blocks\_0\_ffn\_x.pdf}}
\end{subfigure}\hfill
\begin{subfigure}[t]{0.3\textwidth}
\centering
\includegraphics[width=\linewidth]{\detokenize{figures/pca\_ours\_transformer\_blocks\_8\_ffn\_x.pdf}}
\end{subfigure}\hfill
\begin{subfigure}[t]{0.3\textwidth}
\centering
\includegraphics[width=\linewidth]{\detokenize{figures/pca\_ours\_transformer\_blocks\_15\_ffn\_x.pdf}}
\end{subfigure}

\vspace{0.3em}

\begin{subfigure}[t]{0.3\textwidth}
    \centering
    \includegraphics[width=\linewidth]{\detokenize{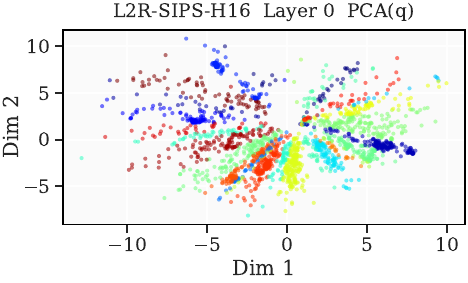}}
\end{subfigure}\hfill
\begin{subfigure}[t]{0.3\textwidth}
    \centering
    \includegraphics[width=\linewidth]{\detokenize{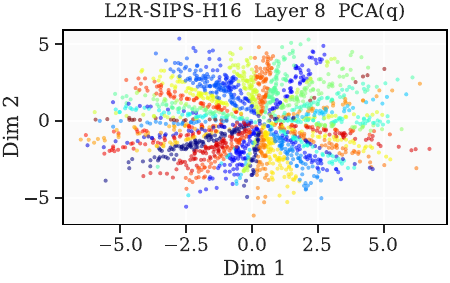}}
\end{subfigure}\hfill
\begin{subfigure}[t]{0.3\textwidth}
    \centering
    \includegraphics[width=\linewidth]{\detokenize{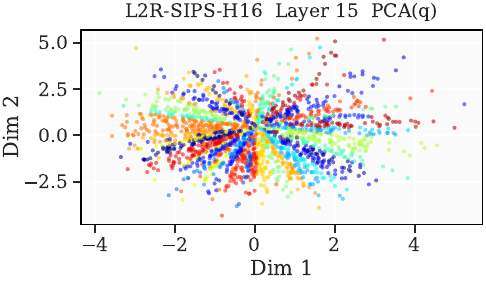}}
\end{subfigure}

\vspace{0.3em}

\begin{subfigure}[t]{0.3\textwidth}
\centering
\includegraphics[width=\linewidth]{\detokenize{figures/pca\_cosine\_transformer\_blocks\_0\_ffn\_x.pdf}}
\end{subfigure}\hfill
\begin{subfigure}[t]{0.3\textwidth}
\centering
\includegraphics[width=\linewidth]{\detokenize{figures/pca\_cosine\_transformer\_blocks\_8\_ffn\_x.pdf}}
\end{subfigure}\hfill
\begin{subfigure}[t]{0.3\textwidth}
\centering
\includegraphics[width=\linewidth]{\detokenize{figures/pca\_cosine\_transformer\_blocks\_15\_ffn\_x.pdf}}
\end{subfigure}

\vspace{0.3em}

\begin{subfigure}[t]{0.3\textwidth}
    \centering
    \includegraphics[width=\linewidth]{\detokenize{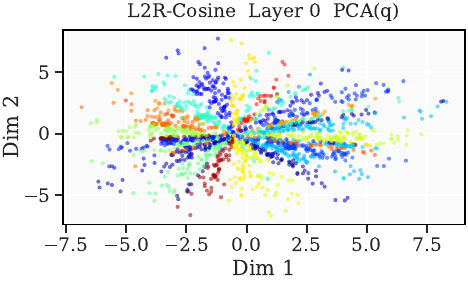}}
\end{subfigure}\hfill
\begin{subfigure}[t]{0.3\textwidth}
    \centering
    \includegraphics[width=\linewidth]{\detokenize{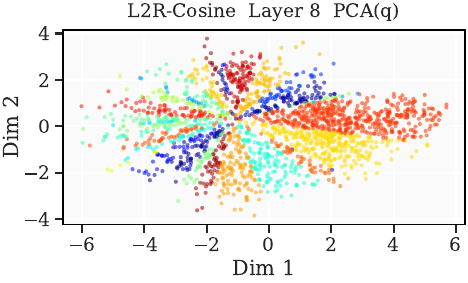}}
\end{subfigure}\hfill
\begin{subfigure}[t]{0.3\textwidth}
    \centering
    \includegraphics[width=\linewidth]{\detokenize{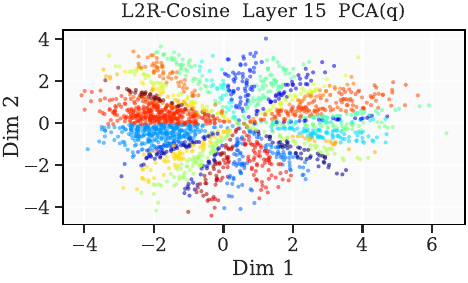}}
\end{subfigure}

\vspace{0.3em}

\begin{subfigure}[t]{0.3\textwidth}
\centering
\includegraphics[width=\linewidth]{\detokenize{figures/pca\_dot\_transformer\_blocks\_0\_ffn\_x.pdf}}
\end{subfigure}\hfill
\begin{subfigure}[t]{0.3\textwidth}
\centering
\includegraphics[width=\linewidth]{\detokenize{figures/pca\_dot\_transformer\_blocks\_8\_ffn\_x.pdf}}
\end{subfigure}\hfill
\begin{subfigure}[t]{0.3\textwidth}
\centering
\includegraphics[width=\linewidth]{\detokenize{figures/pca\_dot\_transformer\_blocks\_15\_ffn\_x.pdf}}
\end{subfigure}

\vspace{0.3em}

\begin{subfigure}[t]{0.3\textwidth}
    \centering
    \includegraphics[width=\linewidth]{\detokenize{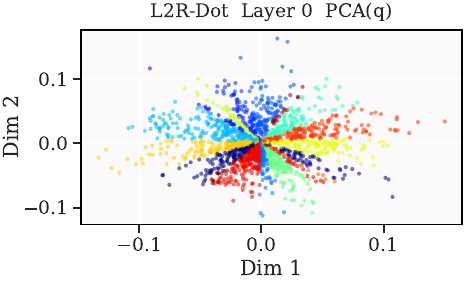}}
\end{subfigure}\hfill
\begin{subfigure}[t]{0.3\textwidth}
    \centering
    \includegraphics[width=\linewidth]{\detokenize{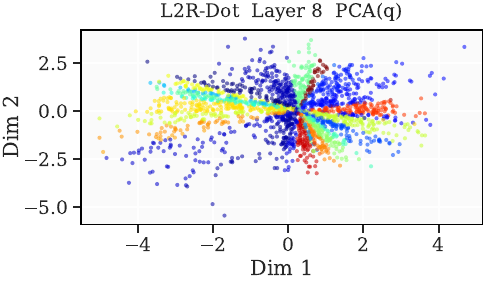}}
\end{subfigure}\hfill
\begin{subfigure}[t]{0.3\textwidth}
    \centering
    \includegraphics[width=\linewidth]{\detokenize{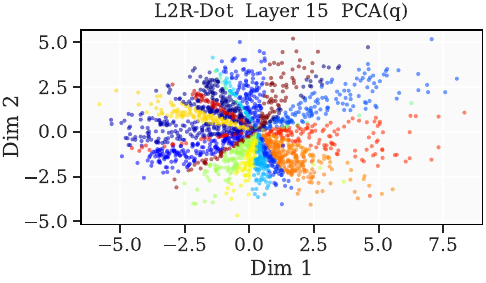}}
\end{subfigure}

\caption{\textbf{Routing geometry under L2R.}
PCA scatter plots for three representative MoE FFN layers in OLMoE.
Points are token representations colored by the top-1 expert.
From top to bottom, the rows show backbone representations ($x$-space)
and routing latent queries ($q$-space) for L2R-SIPS, L2R-Cosine,
and L2R-Dot, respectively.
Compared to $x$-space, the $q$-space geometry typically exhibits
clearer expert-aligned separation and reduced selection ambiguity.}
\label{fig:app_l2r_geometry}

\end{figure}

\clearpage
\begin{figure*}[htbp]
    \centering
    \begin{subfigure}[t]{0.3\textwidth}
        \centering
        \includegraphics[width=\linewidth]{\detokenize{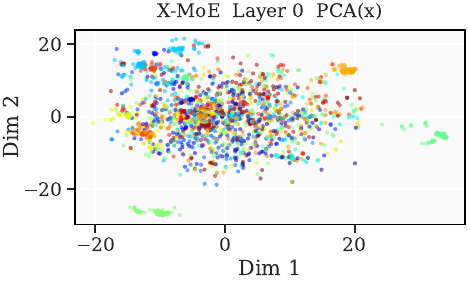}}
    \end{subfigure}
\hfill
    \begin{subfigure}[t]{0.3\textwidth}
        \centering
        \includegraphics[width=\linewidth]{\detokenize{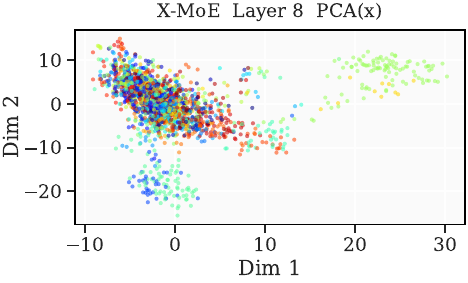}}
    \end{subfigure}
\hfill
    \begin{subfigure}[t]{0.3\textwidth}
        \centering
        \includegraphics[width=\linewidth]{\detokenize{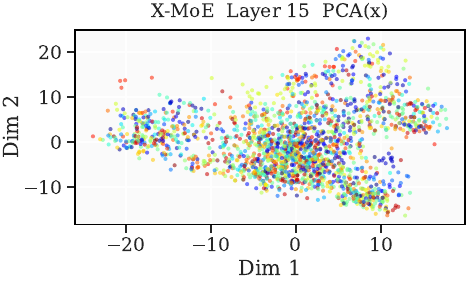}}
    \end{subfigure}
    \vspace{0.3em}
        \begin{subfigure}[t]{0.3\textwidth}
        \centering
        \includegraphics[width=\linewidth]{\detokenize{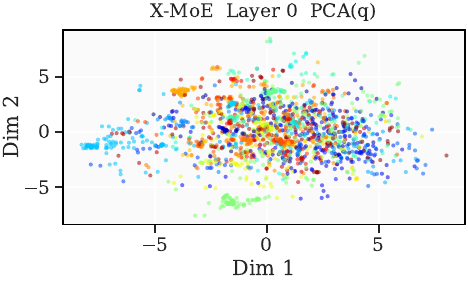}}
    \end{subfigure}
\hfill
    \begin{subfigure}[t]{0.3\textwidth}
        \centering
        \includegraphics[width=\linewidth]{\detokenize{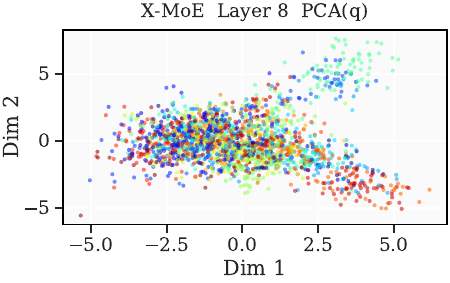}}
    \end{subfigure}
\hfill
    \begin{subfigure}[t]{0.3\textwidth}
        \centering
        \includegraphics[width=\linewidth]{\detokenize{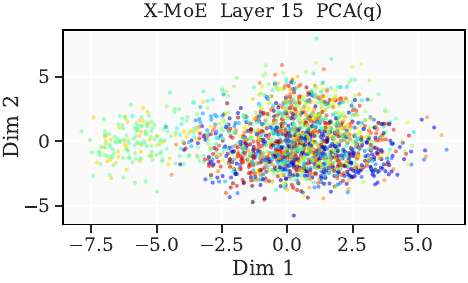}}
    \end{subfigure}

\caption{\textbf{Routing geometry under the X-MoE router.}
PCA projections of token representations at three FFN routing sites (blocks 0/8/15; left to right).
Top: backbone activations in the raw representation space $x$.
Bottom: router features $q$ used for gating at the corresponding blocks.
Each point denotes a token from the same evaluation batch (aggregated across distributed ranks), and colors indicate the selected top-1 expert.
The visualization illustrates how token-to-expert partitioning evolves with depth and how separable the routed groups appear in $x$ versus $q$.}

    \label{fig:app_xmoe_geometry}
\end{figure*}

\subsection{Expert Usage Heatmaps}
\label{sec:appendix_usage}

\begin{figure*}[htbp]
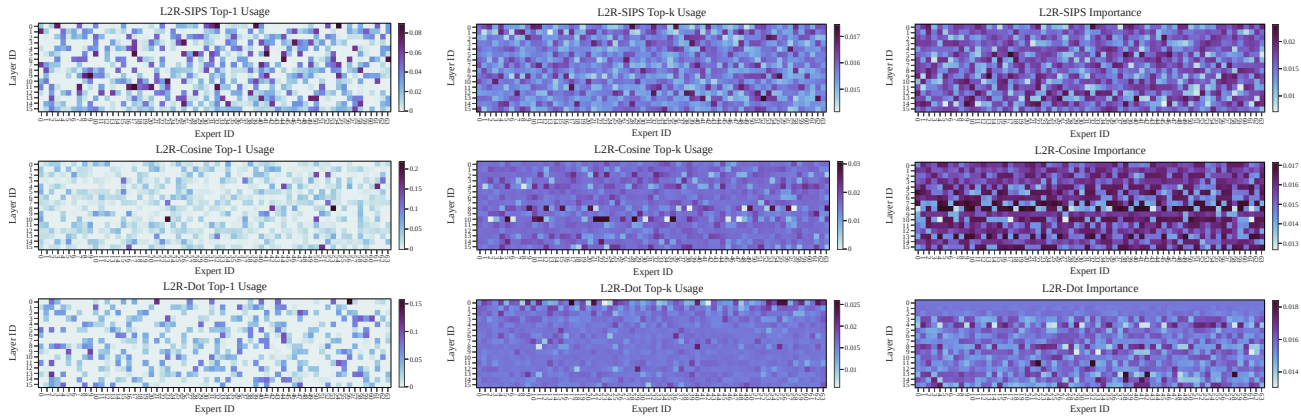

    \centering
    \begin{subfigure}[t]{0.32\textwidth}
        \includegraphics[width=\linewidth]{figures/usage/ours__load_heatmap__top1__plotly.pdf}
    \end{subfigure}
    \hfill
    \begin{subfigure}[t]{0.32\textwidth}
        \includegraphics[width=\linewidth]{figures/usage/ours__load_heatmap__topk__plotly.pdf}
    \end{subfigure}
    \hfill
    \begin{subfigure}[t]{0.32\textwidth}
        \includegraphics[width=\linewidth]{figures/usage/ours__load_heatmap__importance__plotly.pdf}
    \end{subfigure}
    
    \begin{subfigure}[t]{0.32\textwidth}
        \includegraphics[width=\linewidth]{figures/usage/cosine__load_heatmap__top1__plotly.pdf}
    \end{subfigure}
    \hfill
    \begin{subfigure}[t]{0.32\textwidth}
        \includegraphics[width=\linewidth]{figures/usage/cosine__load_heatmap__topk__plotly.pdf}
    \end{subfigure}
    \hfill
    \begin{subfigure}[t]{0.32\textwidth}
        \includegraphics[width=\linewidth]{figures/usage/cosine__load_heatmap__importance__plotly.pdf}
    \end{subfigure}

        \centering
    \begin{subfigure}[t]{0.32\textwidth}
        \includegraphics[width=\linewidth]{figures/usage/dot__load_heatmap__top1__plotly.pdf}
    \end{subfigure}
    \hfill
    \begin{subfigure}[t]{0.32\textwidth}
        \includegraphics[width=\linewidth]{figures/usage/dot__load_heatmap__topk__plotly.pdf}
    \end{subfigure}
    \hfill
    \begin{subfigure}[t]{0.32\textwidth}
        \includegraphics[width=\linewidth]{figures/usage/dot__load_heatmap__importance__plotly.pdf}
    \end{subfigure}

    \caption{\textbf{Expert usage heatmaps for L2R-Cosine, L2R-Dot and L2R-SIPS.}
    From left to right: Top-1 routing frequency, Top-k routing frequency, and importance-based routing weights.
    Each row corresponds to a transformer layer and each column corresponds to an expert.}
    \label{fig:usage_cosine}
    \vspace{-4mm}
\end{figure*}

\begin{figure*}[htbp]
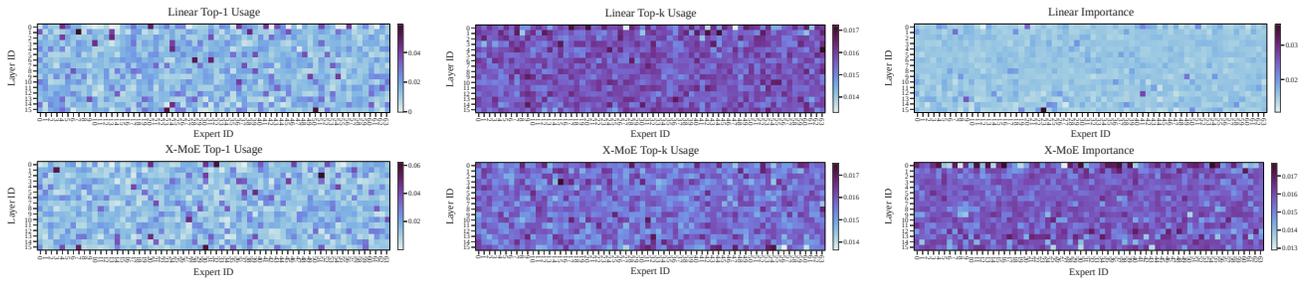

    \centering
    \begin{subfigure}[t]{0.32\textwidth}
        \includegraphics[width=\linewidth]{figures/usage/linear__load_heatmap__top1__plotly.pdf}
    \end{subfigure}
    \hfill
    \begin{subfigure}[t]{0.32\textwidth}
        \includegraphics[width=\linewidth]{figures/usage/linear__load_heatmap__topk__plotly.pdf}
    \end{subfigure}
    \hfill
    \begin{subfigure}[t]{0.32\textwidth}
        \includegraphics[width=\linewidth]{figures/usage/linear__load_heatmap__importance__plotly.pdf}
    \end{subfigure}

        \centering
    \begin{subfigure}[t]{0.32\textwidth}
        \includegraphics[width=\linewidth]{figures/usage/xmoe__load_heatmap__top1__plotly.pdf}
    \end{subfigure}
    \hfill
    \begin{subfigure}[t]{0.32\textwidth}
        \includegraphics[width=\linewidth]{figures/usage/xmoe__load_heatmap__topk__plotly.pdf}
    \end{subfigure}
    \hfill
    \begin{subfigure}[t]{0.32\textwidth}
        \includegraphics[width=\linewidth]{figures/usage/xmoe__load_heatmap__importance__plotly.pdf}
    \end{subfigure}

    \caption{\textbf{Expert usage heatmaps for Linear routing and X-MoE.} From left to right: Top-1 routing frequency, Top-k routing frequency, and importance-based routing weights.
    Each row corresponds to a transformer layer and each column corresponds to an expert.}
    \label{fig:usage_linear}    \vspace{-4mm}
\end{figure*}

\begin{figure*}[htbp]
    \begin{subfigure}[t]{0.32\textwidth}
        \includegraphics[width=\linewidth]{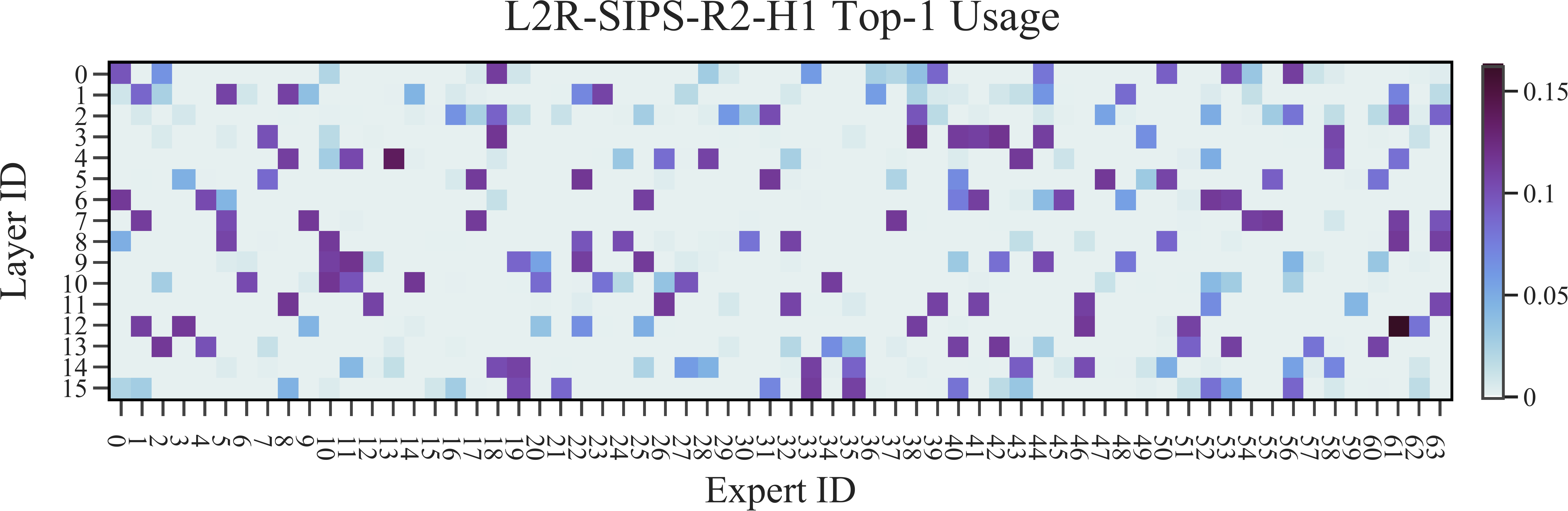}
    \end{subfigure}
    \hfill
    \begin{subfigure}[t]{0.32\textwidth}
        \includegraphics[width=\linewidth]{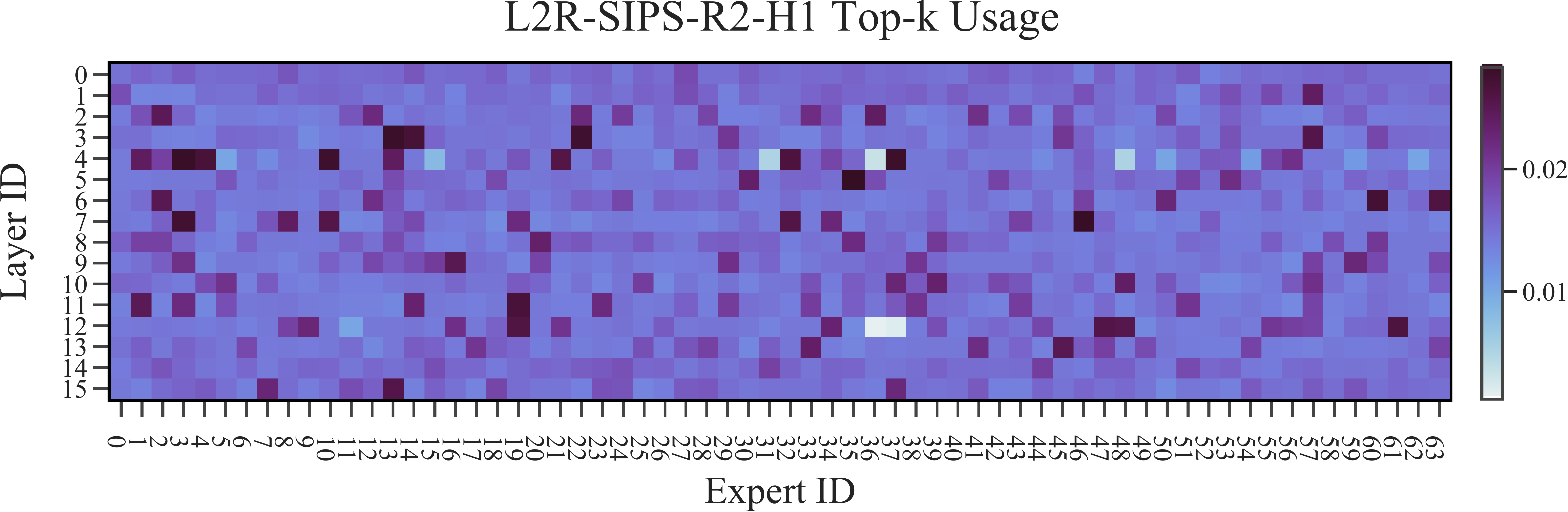}
    \end{subfigure}
    \hfill
    \begin{subfigure}[t]{0.32\textwidth}
        \includegraphics[width=\linewidth]{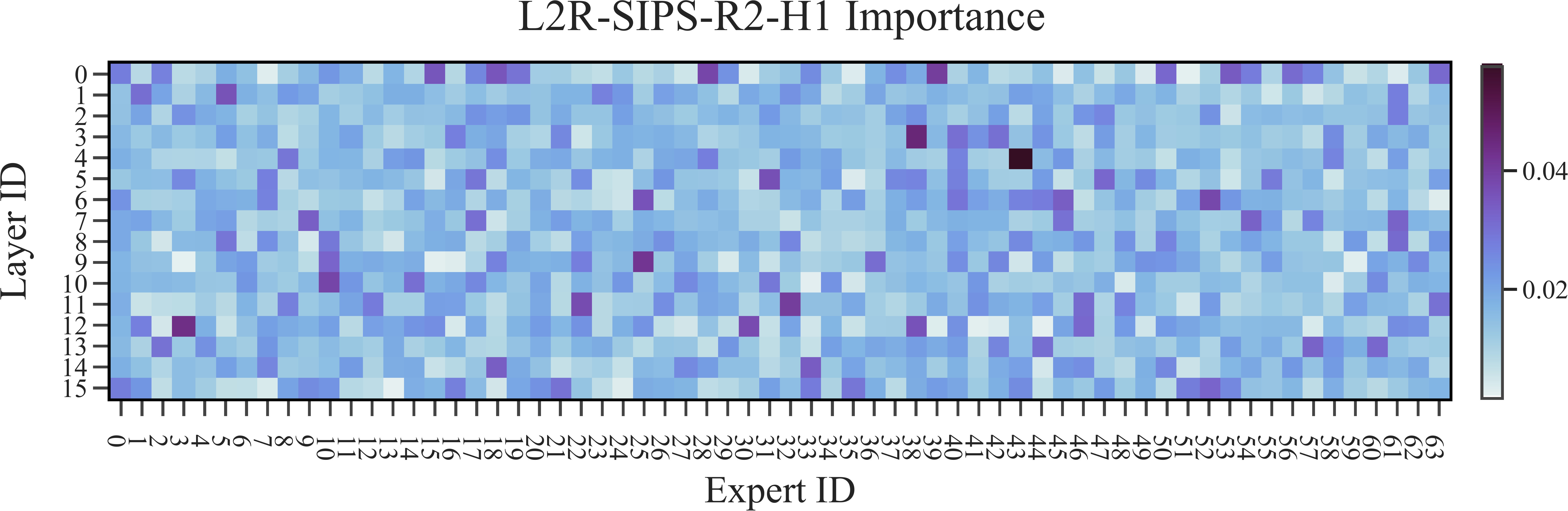}
    \end{subfigure}

    \begin{subfigure}[t]{0.32\textwidth}
        \includegraphics[width=\linewidth]{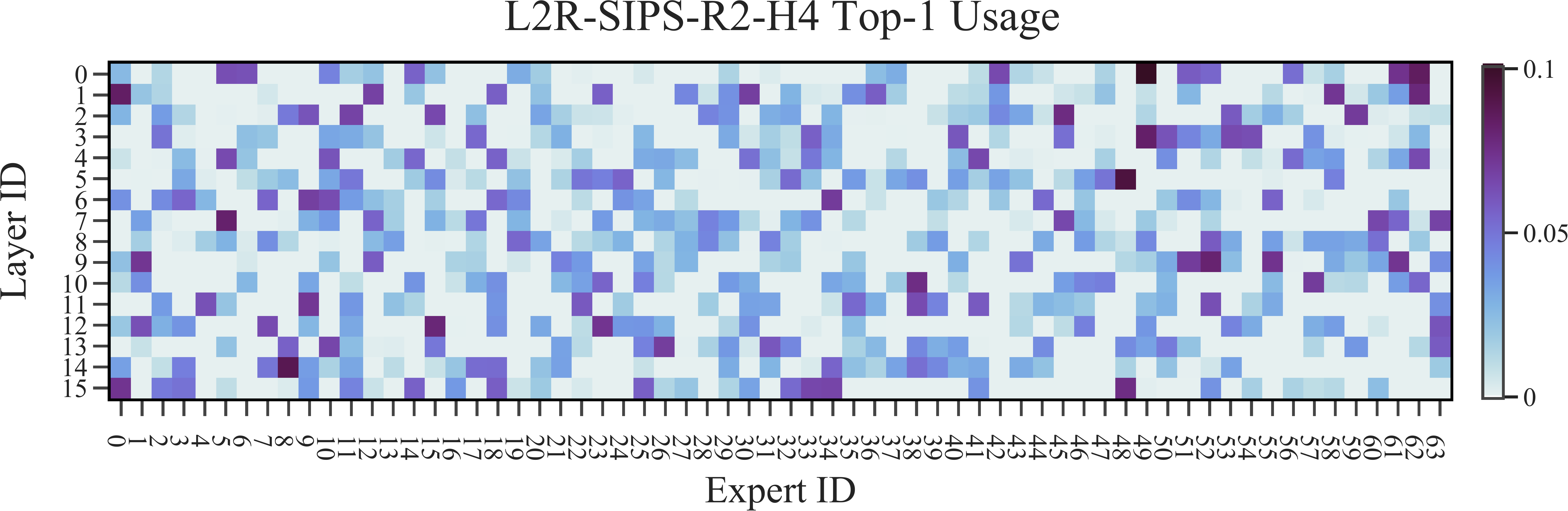}
    \end{subfigure}
    \hfill
    \begin{subfigure}[t]{0.32\textwidth}
        \includegraphics[width=\linewidth]{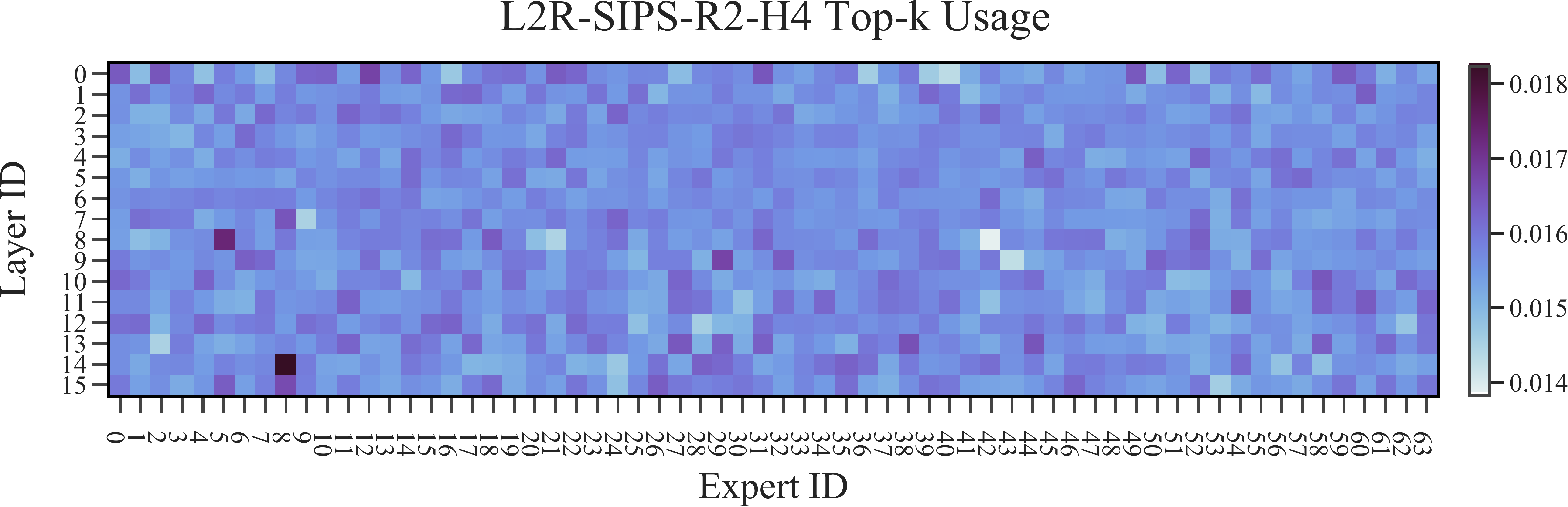}
    \end{subfigure}
    \hfill
    \begin{subfigure}[t]{0.32\textwidth}
        \includegraphics[width=\linewidth]{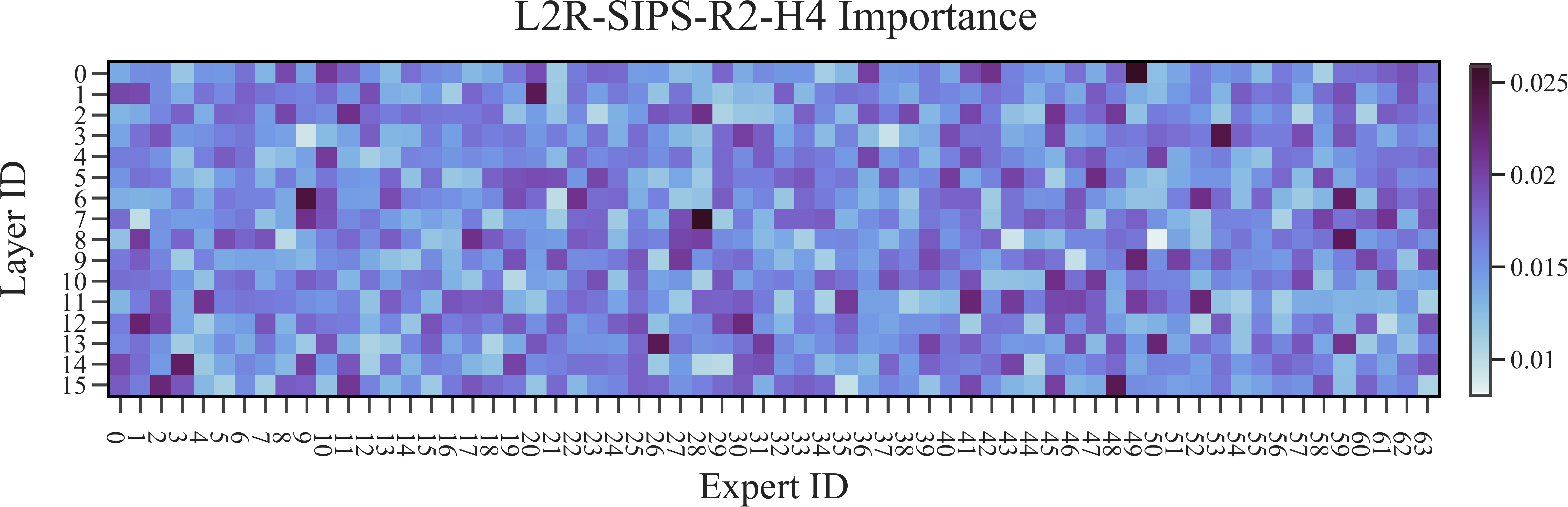}
    \end{subfigure}

       \begin{subfigure}[t]{0.32\textwidth}
        \includegraphics[width=\linewidth]{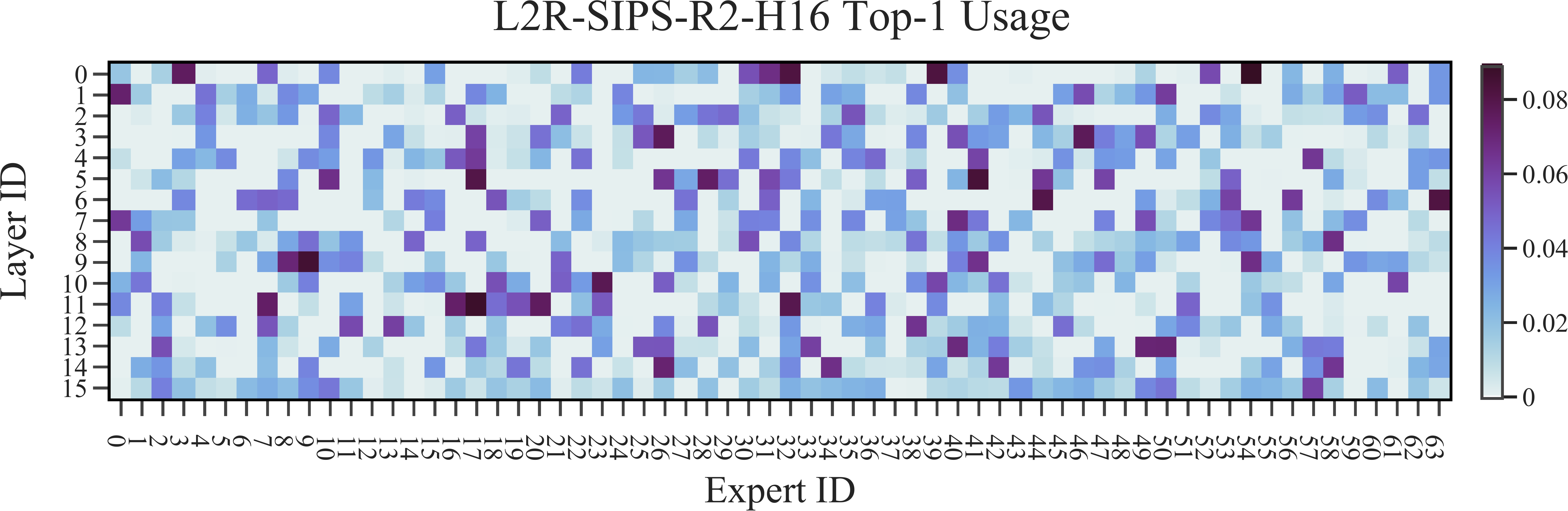}
    \end{subfigure}
    \hfill
    \begin{subfigure}[t]{0.32\textwidth}
        \includegraphics[width=\linewidth]{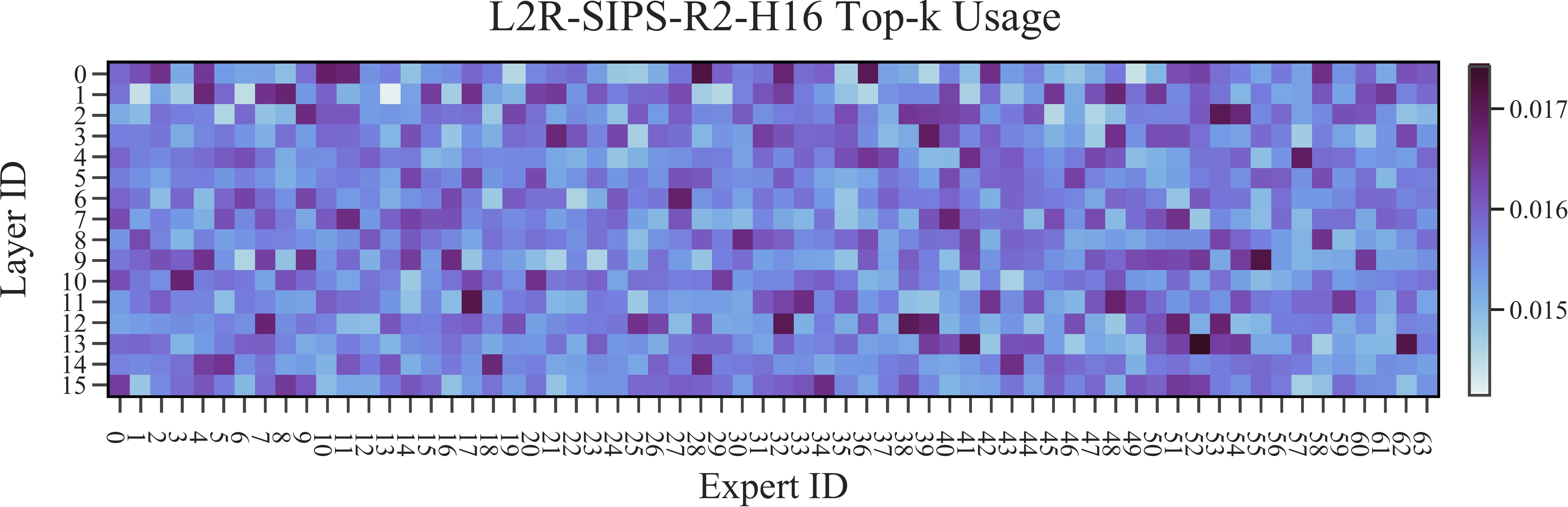}
    \end{subfigure}
    \hfill
    \begin{subfigure}[t]{0.32\textwidth}
        \includegraphics[width=\linewidth]{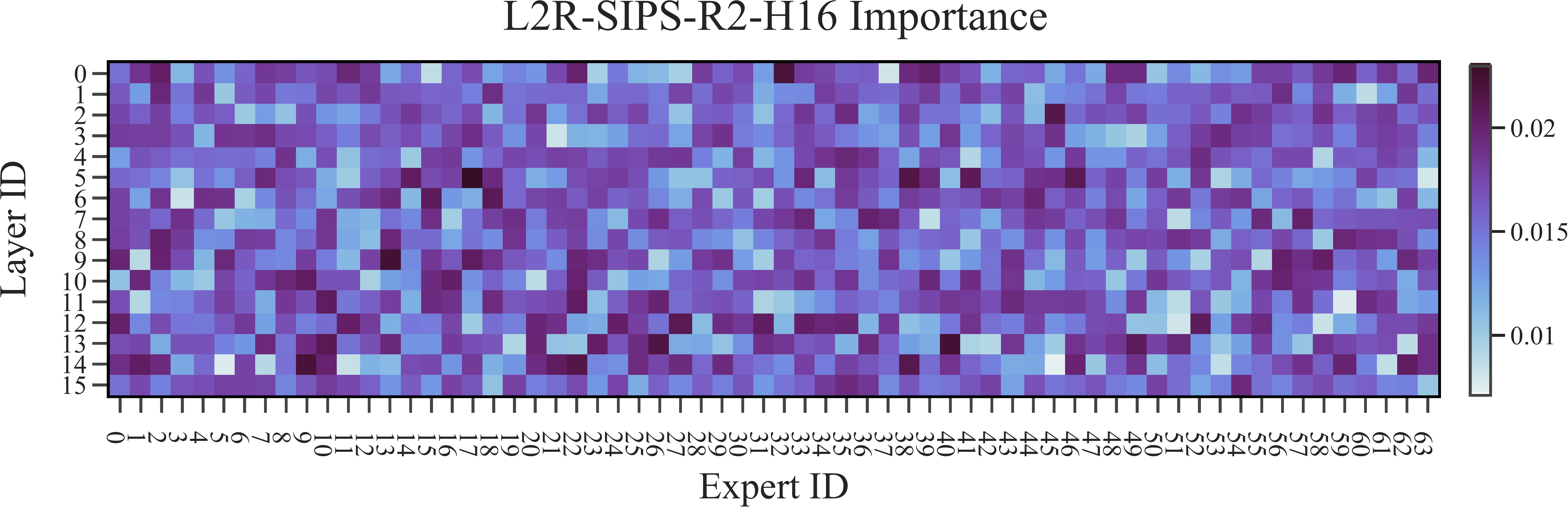}
    \end{subfigure}

    \caption{\textbf{Expert usage heatmaps under different head settings in L2R-SIPS (Rank=2).} From left to right: Top-1 routing frequency, Top-k routing frequency, and importance-based routing weights.
    Each row corresponds to a transformer layer, and each column corresponds to an expert.}
    \label{fig:usage_sips_head_top1}
\end{figure*}

\begin{figure*}[htbp]
    \begin{subfigure}[t]{0.32\textwidth}
        \includegraphics[width=\linewidth]{figures/usage/ours-r2-h4__load_heatmap__top1__plotly.pdf}
    \end{subfigure}
    \hfill
    \begin{subfigure}[t]{0.32\textwidth}
        \includegraphics[width=\linewidth]{figures/usage/ours-r2-h4__load_heatmap__topk__plotly.pdf}
    \end{subfigure}
    \hfill
    \begin{subfigure}[t]{0.32\textwidth}
        \includegraphics[width=\linewidth]{figures/usage/ours-r2-h4__load_heatmap__importance__plotly.pdf}
    \end{subfigure}

    \begin{subfigure}[t]{0.32\textwidth}
        \includegraphics[width=\linewidth]{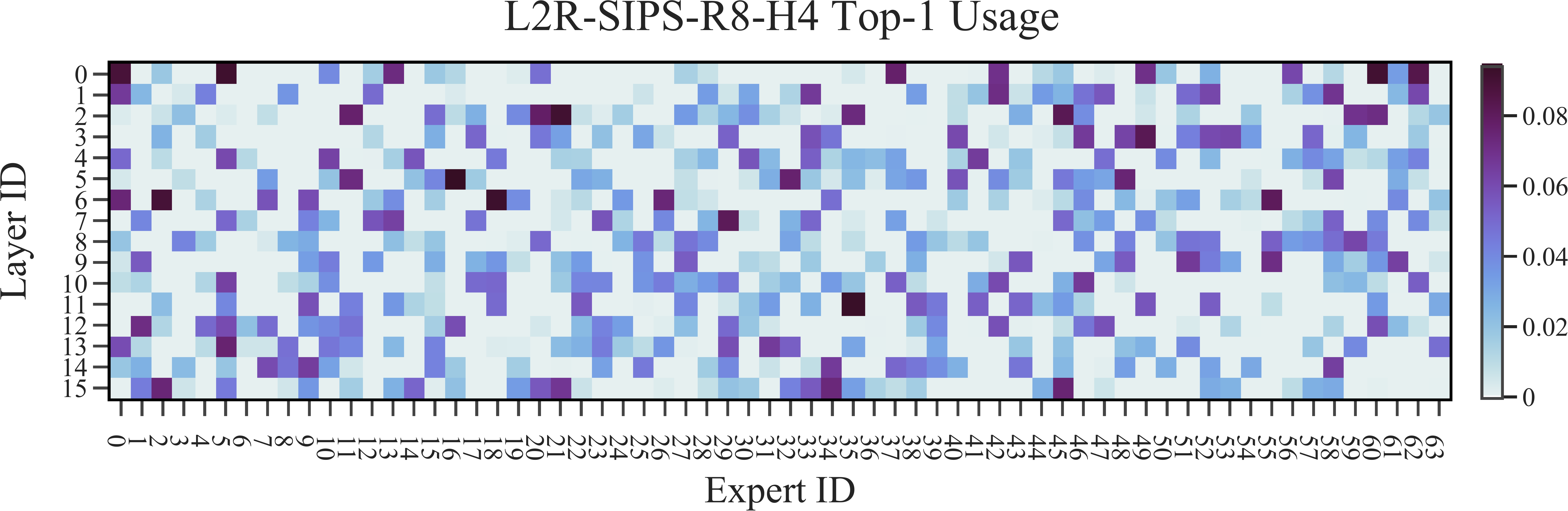}
    \end{subfigure}
    \hfill
    \begin{subfigure}[t]{0.32\textwidth}
        \includegraphics[width=\linewidth]{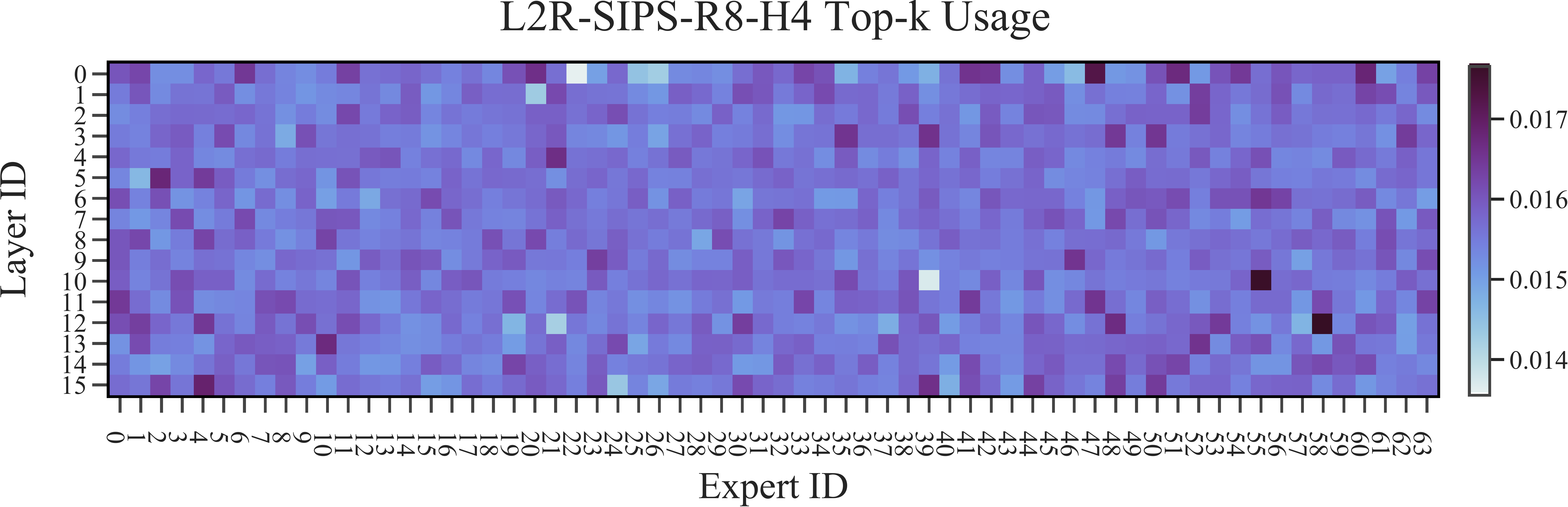}
    \end{subfigure}
    \hfill
    \begin{subfigure}[t]{0.32\textwidth}
        \includegraphics[width=\linewidth]{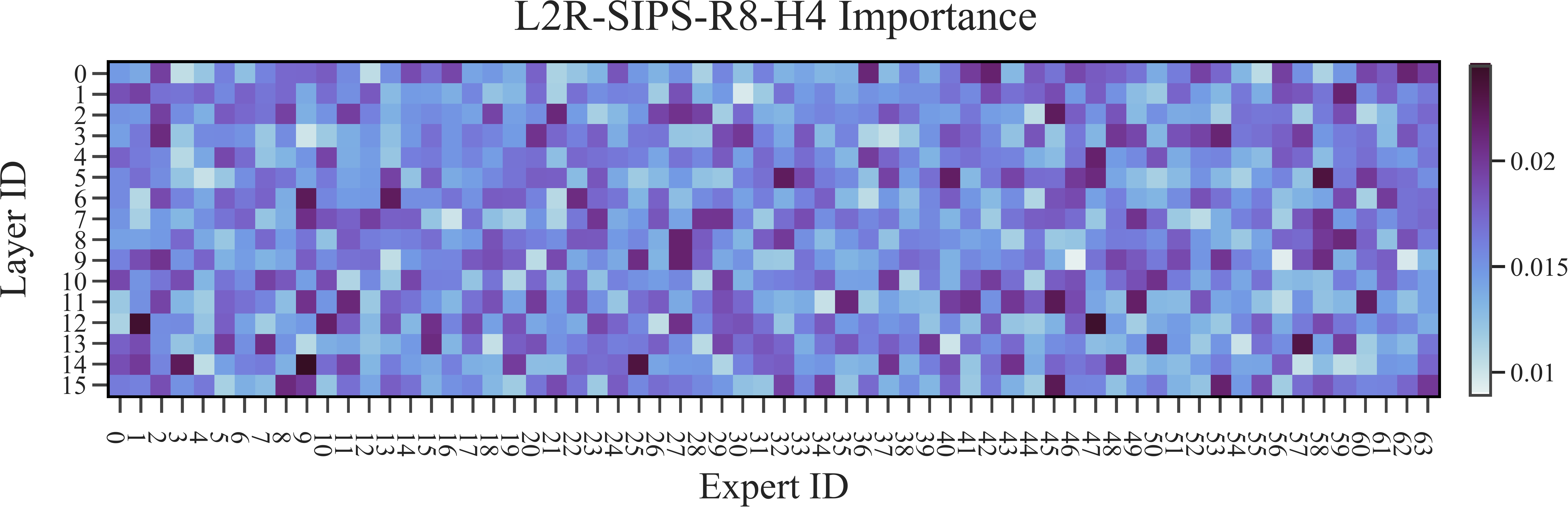}
    \end{subfigure}

    \begin{subfigure}[t]{0.32\textwidth}
        \includegraphics[width=\linewidth]{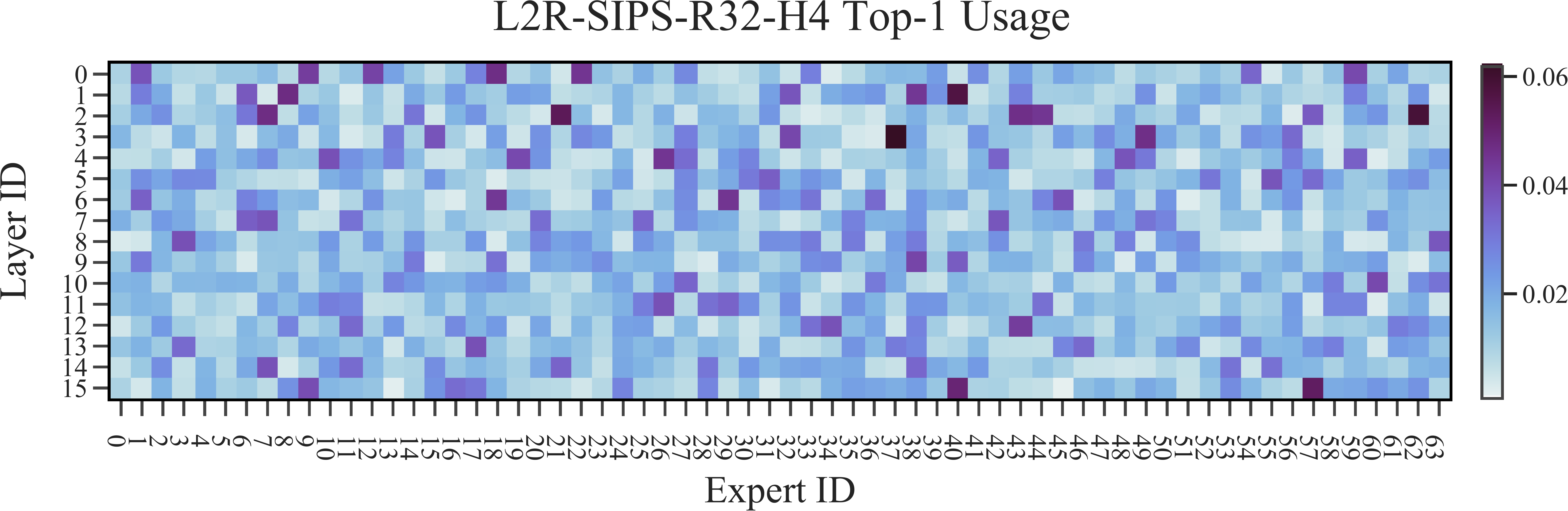}
    \end{subfigure}
    \hfill
    \begin{subfigure}[t]{0.32\textwidth}
        \includegraphics[width=\linewidth]{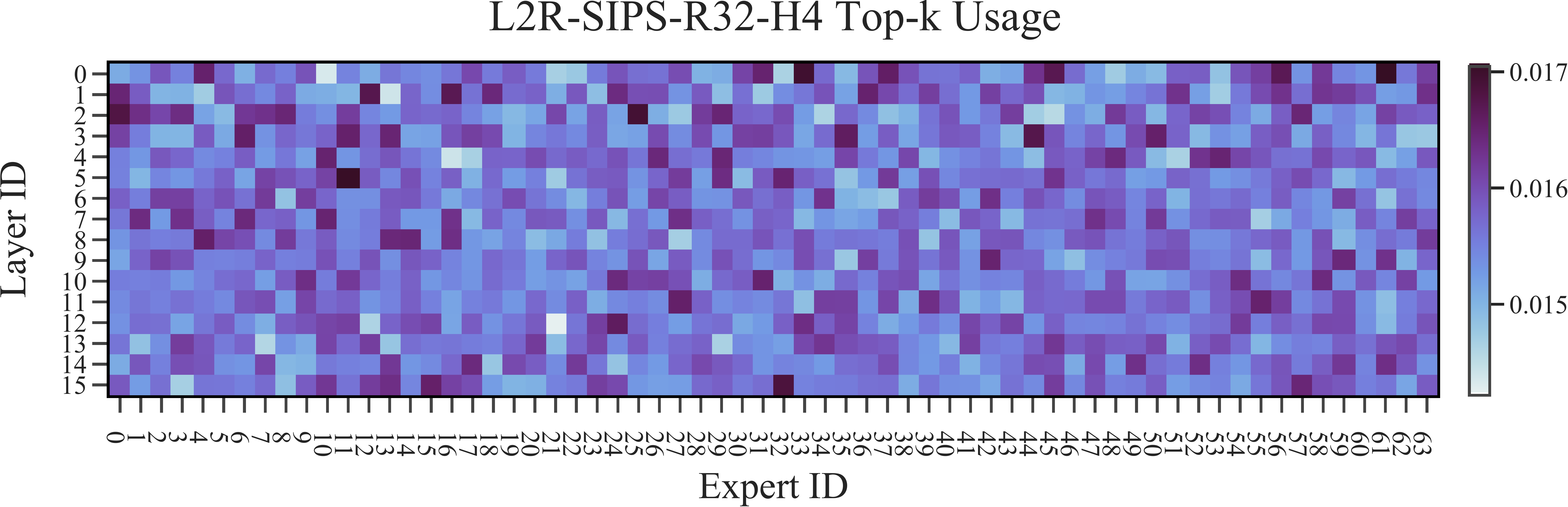}
    \end{subfigure}
    \hfill
    \begin{subfigure}[t]{0.32\textwidth}
        \includegraphics[width=\linewidth]{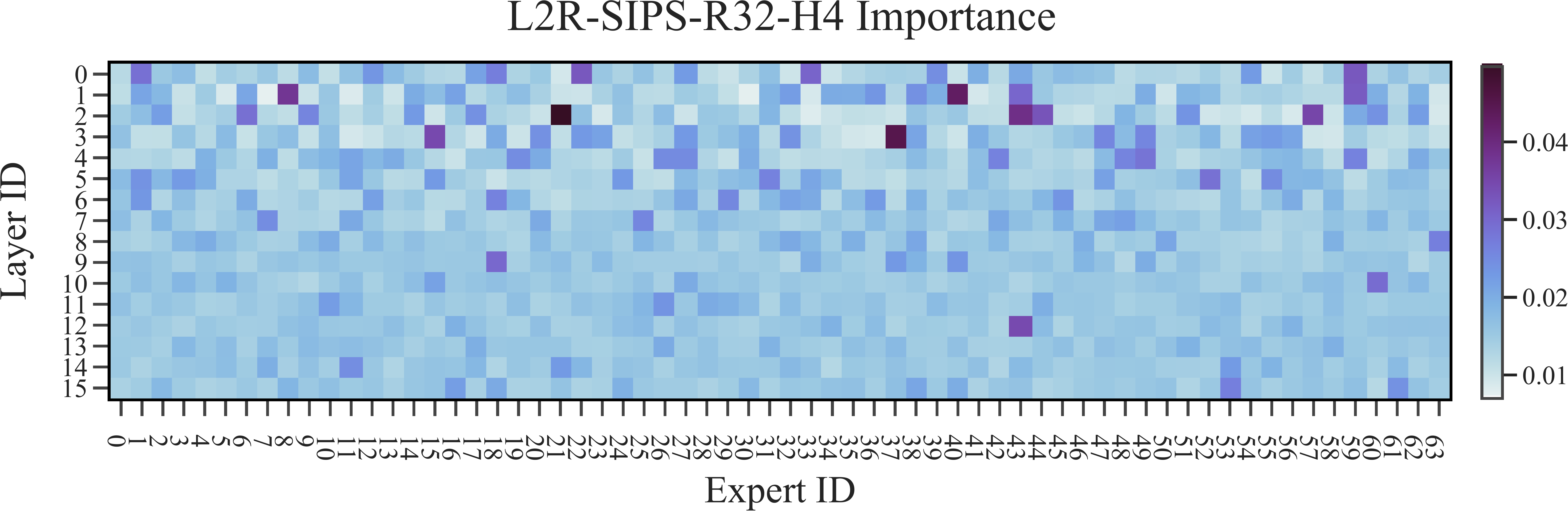}
    \end{subfigure}

    \caption{\textbf{Expert usage heatmaps under different rank settings in L2R-SIPS (Head=4).} From left to right: Top-1 routing frequency, Top-k routing frequency, and importance-based routing weights.
    Each row corresponds to a transformer layer, and each column corresponds to an expert.}
    \label{fig:usage_sips_rank_top1}
\end{figure*}

\newpage

\end{document}